\documentclass[runningheads]{llncs}
\usepackage{graphicx}

\usepackage{tikz}
\usepackage{comment}
\usepackage{amsmath,amssymb} 
\usepackage{color}
\usepackage{pifont}
\usepackage{fontawesome5}
\usepackage{multirow}
\usepackage{cite}
\usepackage{soul}
\usepackage[abs]{overpic}
\usepackage{xspace}
\usepackage{rotating}
\usepackage{booktabs}
\usepackage{arydshln}
\usepackage{colortbl}
\usepackage{xcolor}
\usepackage{bbm}

\usepackage[pagebackref=true,breaklinks=true,letterpaper=true,colorlinks,bookmarks=false]{hyperref}

\usepackage{microtype}

\usepackage[accsupp]{axessibility}  

\usepackage[utf8]{inputenc}
\usepackage{enumitem}

\newcommand{\ourmethod}{GIPSO\xspace}
\newcommand{\CC}[1]{\cellcolor{#1}}
\definecolor{sourcecolor}{rgb}{0.78, 0.78, 0.78}
\definecolor{bestcolor}{rgb}{0.5, 0.95, 0.5}

\begin{document}
\pagestyle{headings}
\mainmatter
\def\ECCVSubNumber{2016}  

\title{GIPSO: 
Geometrically Informed Propagation for 
Online Adaptation in 3D LiDAR Segmentation} 


\titlerunning{Abbreviated paper title}
%
\titlerunning{GIPSO: Geometrically Informed Prop. for Online Adapt. in 3D LiDAR Seg.}
%

\author{Cristiano Saltori\inst{1} \and
Evgeny Krivosheev \inst{1} \and
Stéphane Lathuilière\inst{2} \and
Nicu Sebe\inst{1} \and
Fabio Galasso\inst{3} \and
Giuseppe Fiameni\inst{4} \and
Elisa Ricci\inst{1,5} \and
Fabio Poiesi\inst{5}
}

%
\authorrunning{C. Saltori et al.}
%
\institute{University of Trento, Trento, Italy \and
LTCI, Télécom-Paris, Intitute Polytechnique de Paris, Palaiseau, France \and
Sapienza University of Rome, Rome, Italy \and
NVIDIA AI Technology Center \and
Fondazione Bruno Kessler, Trento, Italy\\
\email{cristiano.saltori@unitn.it}}
\maketitle

\begin{abstract}
   3D point cloud semantic segmentation is fundamental for autonomous driving.
   Most approaches in the literature neglect an important aspect, i.e., how to deal with domain shift when handling dynamic scenes.
   This can significantly hinder the navigation capabilities of self-driving vehicles. 
   This paper advances the state of the art in this research field.
   Our first contribution consists in analysing a new unexplored scenario in point cloud segmentation, namely Source-Free Online Unsupervised Domain Adaptation (SF-OUDA).
   We experimentally show that state-of-the-art methods have a rather limited ability to adapt pre-trained deep network models to unseen domains in an online manner.
   Our second contribution is an approach that relies on adaptive self-training and geometric-feature propagation to adapt a pre-trained source model online without requiring either source data or target labels.
   Our third contribution is to study SF-OUDA in a challenging setup where source data is synthetic and target data is point clouds captured in the real world.
   We use the recent SynLiDAR dataset as a synthetic source and introduce two new synthetic (source) datasets, which can stimulate future synthetic-to-real autonomous driving research.
   Our experiments show the effectiveness of our segmentation approach on thousands of real-world point clouds. Code and synthetic datasets are available at \url{https://github.com/saltoricristiano/gipso-sfouda}.
   
   
\end{abstract}

\section{Introduction}\label{sec:intro}
\vspace{-.2cm}
Autonomous driving requires accurate and efficient 3D visual scene perception algorithms. 
Low-level visual tasks such as detection and segmentation are crucial to enable higher-level tasks such as path planning \cite{liu2017path, dolgov2010path} and obstacle avoidance \cite{rosolia2016autonomous}.
Deep learning-based methods have proven to be the most suitable option to meet these requirements so far, but at the cost of requiring large-scale annotated dataset for training \cite{lecun2015deep}. 
Relying only on annotated data is not always a viable solution. 
This problem can be mitigated by considering synthetic data, as it can be generated at low cost with potentially unlimited annotations and under different environmental conditions \cite{Dosovitskiy17, hurl2019precise}.
However, when a model trained on synthetic data is deployed in the real world, typically it will underperform due to domain shift, \textit{e.g.}, caused by varying lighting conditions, clutter, 
occlusions and materials with different reflective properties \cite{torralba2011unbiased}.
\begin{figure}[t]
\centering
\includegraphics[width=1\columnwidth]{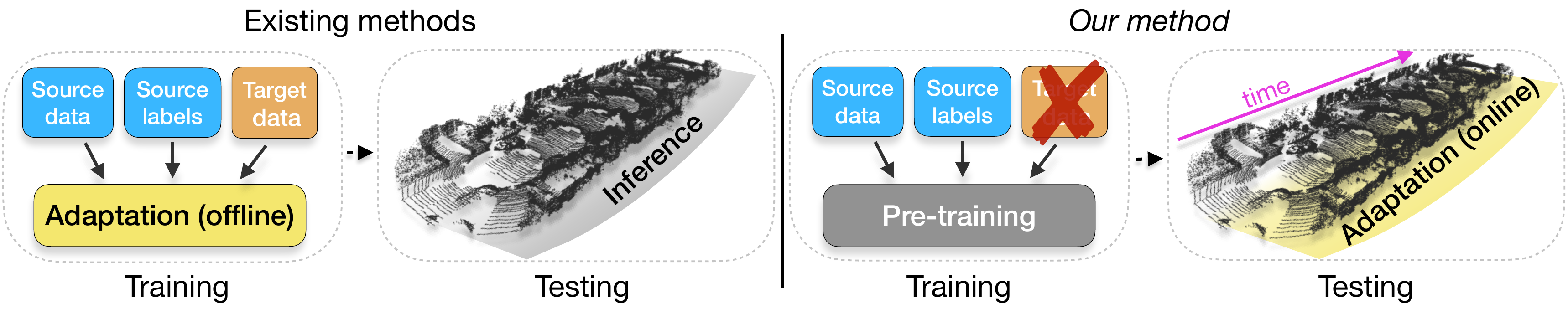}
\vspace{-.6cm}
\caption{Existing methods adapt 3D semantic segmentation networks \textit{offline}, requiring both source and target data. Differently, real-world applications urge solutions capable of adapting to unseen scenes online having access only to a pre-trained model.}
\label{fig:teaser}
\vspace{-.4cm}
\end{figure}
We argue that a 3D semantic segmentation algorithm running on an autonomous vehicle should be capable of adapting online -- handling scenarios that are visited for the first time while driving -- and it should do so by only using the newly captured data.
A variety of research works have addressed the adaptation problem in the context of 3D semantic segmentation.
However, most approaches operate offline and assume to have access to training (source) data \cite{langer2020domain, zhao2020epointda, yi2021complete, zou2018unsupervised, zou2019confidence, wu2019squeezesegv2}.
In this paper, we argue that these two assumptions are too restrictive in an autonomous driving scenario (Fig.~\ref{fig:teaser}). 
On the one hand, offline adaptation would be equivalent to performing model adaptation on the data a vehicle has captured when the navigation has terminated, which is clearly a sub-optimal solution for autonomous driving \cite{levinson2011towards}.
On the other hand, having to rely on source data may not be a viable option, as it requires the method to store and query potentially large amount of data, thus hindering scalability \cite{liang2020we, liu2021source}.

To overcome these limitations, in this paper we explore the new problem of Source-Free Online Unsupervised Domain Adaptation (SF-OUDA) for semantic segmentation, \textit{i.e.}, that of adapting a deep semantic segmentation model while a vehicle navigates in an unseen environment without relying on human supervision.
Specifically, in this work we first implement, adapt and thoroughly analyze existing adaptation methods for the 3D semantic segmentation problem in a SF-OUDA setup.
We experimentally observe that none of these methods provides consistent and satisfactory performance when employed in a SF-OUDA setting.
However, there are elements of interest that, when carefully combined and extended, can be generally applicable.
This leads us to move toward and design \ourmethod (Geometrically Informed Propagation for Source-free Online adaptation), the first SF-OUDA method for 3D point cloud segmentation that builds upon recent advances in the literature, and exploits geometry information and temporal consistency to support the domain adaptation process.
We also introduce two new synthetic datasets to benchmark SF-OUDA in two different real-world datasets, \textit{i.e.}~SemanticKITTI \cite{behley2019iccv, geiger2012we, geiger2013vision} and nuScenes \cite{nuscenes2019}.
We validate our approach on these new synthetic-to-real benchmarks.
Our motivation for creating these datasets is to make evaluation more comprehensive and to assess the generalization ability of different techniques to different experimental setups. 
In summary, our contributions are:
\setlist{nolistsep}
\begin{itemize}[noitemsep]
\item A thorough experimental analysis of existing domain adaptation methods for 3D semantic segmentation in a SF-OUDA setting;
\item A novel method for SF-OUDA that exploits low-level geometric properties and temporal information to continuously adapt a 3D segmentation model;
\item The introduction of two new LiDAR synthetic datasets that are compatible with the SemanticKITTI and nuScenes datasets.
\end{itemize}

\section{Related work}\label{sec:related}
\vspace{-.2cm}
\noindent\textbf{Point cloud semantic segmentation.}
Point cloud segmentation methods can be classified into quantization-free and quantization-based architectures.
The former processes the input point clouds in their original 3D format.
Examples include PointNet~\cite{qi2017pointnet} that is based on a series of multi layer perceptrons.
PointNet++~\cite{qi2017pointnet++} builds upon PointNet by using multi-scale sampling and neighbourhood aggregation to encode both global and local features.
RandLA-Net~\cite{hu2020randla} extends PoinNet++ \cite{qi2017pointnet++} by embedding local spatial encoding, random sampling and attentive pooling. 
These methods are computationally inefficient when large-scale point clouds are used.
The latter provides a computationally efficient alternative as input point clouds can be mapped into efficient representations, namely range maps~\cite{milioto2019rangenet++, wu2018squeezeseg, wu2019squeezesegv2}, polar maps~\cite{zhang2020polarnet}, 3D voxel grids~\cite{zhou2018voxelnet, SubmanifoldSparseConvNet, 3DSemanticSegmentationWithSubmanifoldSparseConvNet, choy20194d} or 3D cylindrical voxels~\cite{zhu2021cylindrical}.
Quantization-based approaches can be based on sparse convolutions~\cite{3DSemanticSegmentationWithSubmanifoldSparseConvNet, SubmanifoldSparseConvNet} or Minkowski convolutions~\cite{choy20194d}.
We use the Minkowski Engine \cite{choy20194d} as it provides a suitable trade off between accuracy and efficiency.

\noindent\textbf{Unsupervised domain adaptation.}
Offline UDA can be performed either using source data~\cite{hoffman2018cycada, long2018conditional, saito2018maximum, zou2018unsupervised} or without using source data (source-free UDA)~\cite{liu2021source, liang2020we, saltori2020sf, yang2021exploiting}. 
Online UDA can be used to adapt a model to an unlabelled continuous target data stream through source domain supervision~\cite{volpi2022road}.
It can be employed for classification \cite{moon2020multi}, image semantic segmentation~\cite{volpi2022road}, depth estimation \cite{tonioni2019real, zhang2019online}, robot manipulation \cite{mancini2018kitting}, human mesh reconstruction \cite{Guan_2021_CVPR} and occupancy mapping \cite{tompkins2020online}. 
The assumption of unsupervised target input data can be relaxed and applied for online adaptation in classification \cite{li2020online}, video-object segmentation \cite{voigtlaender2017online} and motion planning \cite{8246875}. 
Recently, test-time adaptation methods have been applied to online UDA in classification by using supervision from source data~\cite{sun2020test, schneider2020improving, wang2021tent}.
We tackle source-free online UDA for point cloud segmentation for the first time.


\noindent\textbf{Domain adaptation for point cloud segmentation.}
Domain shift in point cloud segmentation occurs due to differences in (i) sampling noise, (ii) structure of the environment and (iii) class distributions \cite{jaritz2019xmuda, wu2019squeezesegv2, yi2021complete, zhao2020epointda}.
The domain adaptation problem can be formulated as a 3D surface completion task \cite{yi2021complete} or addressed with ray casting system capable of transferring the target sensor sampling pattern to the source data \cite{langer2020domain}.
Other approaches tackle the domain adaptation problem in the synthetic-to-real setting (\textit{i.e.,} point cloud in the source domain are synthetic, while target ones are collected with LiDAR sensors) \cite{wu2018squeezeseg, wu2019squeezesegv2, zhao2020epointda}.
Attention models can be used to aggregate contextual information with large receptive fields at early layers of the model \cite{wu2018squeezeseg,wu2019squeezesegv2}. 
Geodesic correlation alignment and progressive domain calibration can be also used to further improve domain adaptation effectiveness \cite{wu2019squeezesegv2}.
Authors in \cite{zhao2020epointda} argue that the method in \cite{wu2019squeezesegv2} cannot be trained end-to-end as it employs a multi-stage pipeline. Therefore, they propose an end-to-end approach to simulate the dropout noise of real sensors on synthetic data through a generative adversarial network.
Unlike these methods, we focus on SF-OUDA and propose a novel adaptation method which invokes geometry for propagating reliable pseudo-labels on target data.

\begin{table}[t]
    \tabcolsep 5pt
    \centering\caption{Comparison between public synthetic datasets and Synth4D in terms of sensor specifications, acquisition areas, number of scans, number of points, presence of odometry data, and whether the semantic classes are all or partially shared.}
    \vspace{-.2cm}
    \label{tab:dataset_comparison}
    \resizebox{0.99\columnwidth}{!}{
    \begin{tabular}{l|cccccccc}
        \toprule
        & \multicolumn{2}{c}{Specifications} & \multirow{2}{*}{Areas} & \multirow{2}{*}{Scans} & \multirow{2}{*}{Points} &  \multirow{2}{*}{Odometry} & \multicolumn{2}{c}{Shared semantic classes} \\
        Name & Sensor & FOV & & & & & S-KITTI~[\textcolor{green}{3}] & nuScenes~[\textcolor{green}{4}]\\ 
        \midrule
        SinthCity~\cite{griffiths2019synthcity} & MLS & 360$^{\circ}$ & city & 1 & 367M & & no & no \\
        GTA-LiDAR~\cite{wu2019squeezesegv2} & HDL64E & 90$^{\circ}$ & town & 121087 & - & & partial & no \\
        PreSIL~\cite{hurl2019precise} & HDL64E & 90$^{\circ}$ & town & 51074 & 3135M & & partial & no \\
        \multirow{2}{*}{SynLiDAR~\cite{synlidar}} & \multirow{2}{*}{HDL64E} & \multirow{2}{*}{360$^{\circ}$} & city, town & \multirow{2}{*}{198396} & \multirow{2}{*}{19482M} & & \multirow{2}{*}{all} & \multirow{2}{*}{no}
        \\
         & & & harbor, rural & & & & & \\
        \vspace{-.3cm} & & & & & & & & \\
        \midrule
        \vspace{-.3cm} & & & & & \\
        \multirow{2}{*}{Synth4D (ours)} & HDL64E & \multirow{2}{*}{360$^{\circ}$} & city, town & 20000 & 2000M & \multirow{2}{*}{\ding{51}} & \multirow{2}{*}{all} & \multirow{2}{*}{all} \\
        & HDL32E & & rural, highway  & 20000 & 2000M & & & \\
        \bottomrule
    \end{tabular}}
    \vspace{-0.5cm}
\end{table}

\section{Datasets for synthetic-to-real adaptation}
\vspace{-.2cm}
Autonomous driving simulators enable users to create ad-hoc synthetic datasets that can resemble real-world scenarios.
Examples of popular simulators are GTA-V \cite{Yue2018} and CARLA \cite{Dosovitskiy17}.
In principle, synthetic datasets should be compatible with their real-world counterpart \cite{behley2019iccv, nuscenes2019, geiger2012we}, \textit{i.e.}, they should share the same semantic classes and the same sensor specifications, such as the resolution (32 vs.~64 channels) and the horizontal field of view (e.g., 90$^\circ$ vs.~360$^\circ$).
However, this is not the case for most of the synthetic datasets in literature.
The SynthCity \cite{griffiths2019synthcity} dataset contains large-scale point clouds that are generated from collections of several LiDAR scans, making it unsuitable for online domain adaptation as no odometry data is provided.
PreSIL \cite{hurl2019precise} and GTA-LiDAR's \cite{wu2019squeezesegv2} point clouds are captured from a moving vehicle using a simulated Velodyne HDL64E \cite{velodyne}, as that of SemanticKITTI, however they are rendered with a different field of view, \textit{i.e.}, $90^\circ$ as opposed to $360^\circ$ of SemantiKITTI.
SynLIDAR's~\cite{synlidar} point clouds are obtained using a simulated Velodyne HDL64E with $360^\circ$ field of view, as in SemantiKITTI.
However, the odometry data is not provided, \textit{i.e.}, point clouds are all configured in their local reference frame.
Therefore, domain adaptation algorithms that are based on ray-casting like \cite{langer2020domain} cannot be used.

\begin{figure}[t]
\centering
\begin{tabular}{@{}c@{}c}
    \begin{overpic}[width=.45\columnwidth]{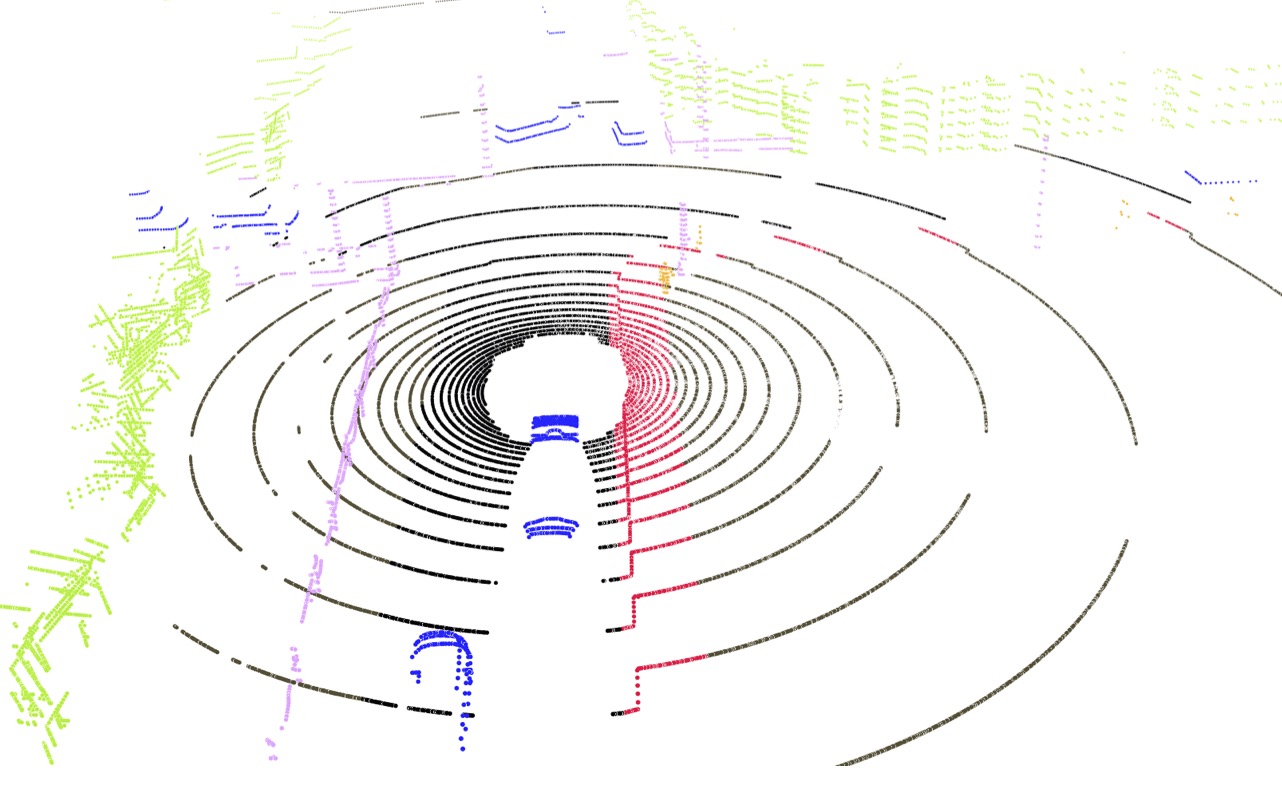}
        \put(63, 0){\color{black}\scriptsize\textbf{(a)}}
    \end{overpic}&
    \hspace{0.5cm}
    \begin{overpic}[width=.45\columnwidth]{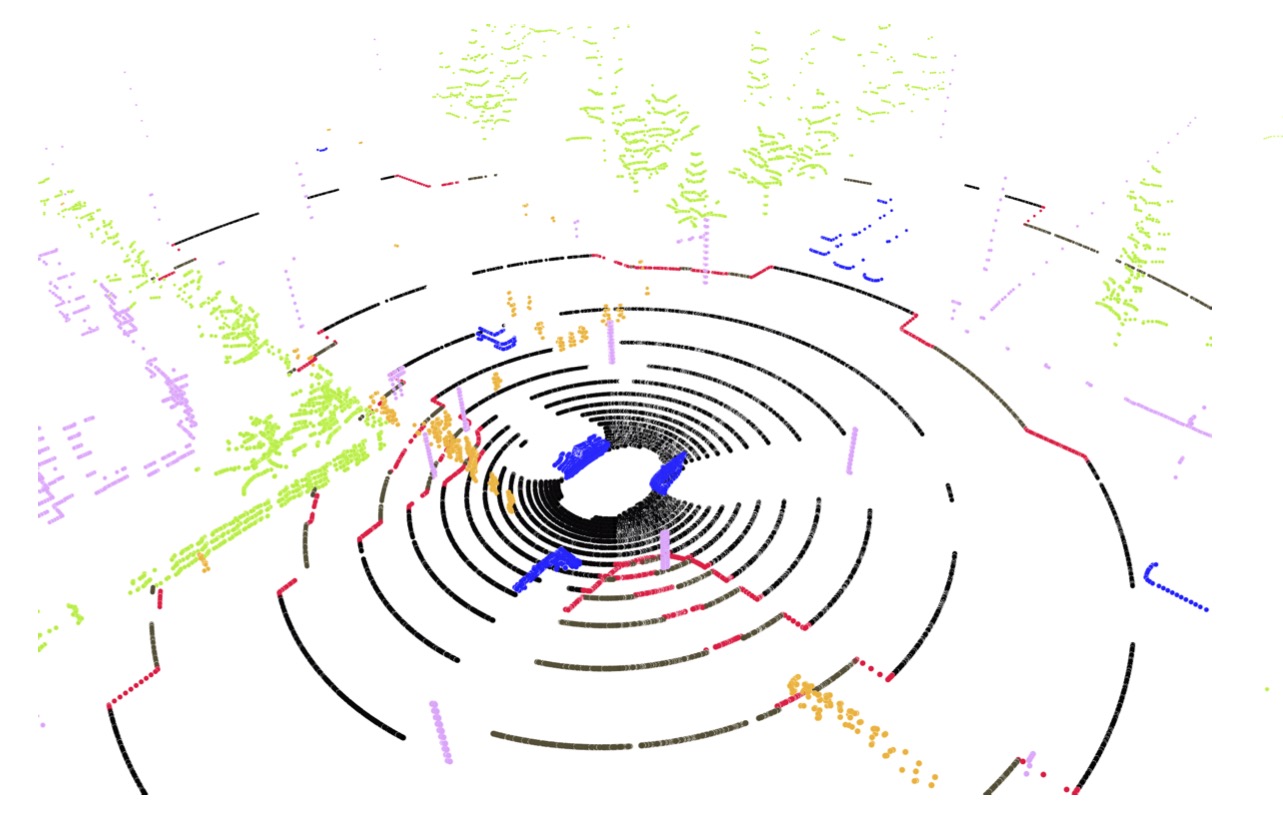}
        \put(67,  0){\color{black}\scriptsize\textbf{(b)}}
    \end{overpic}\\
\end{tabular}
\begin{tabular}{@{}c}
    \begin{overpic}[width=0.99\columnwidth]{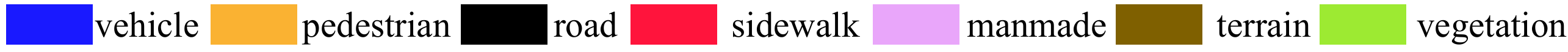}
    \end{overpic}\\
\end{tabular}
\vspace{-.3cm}
\caption{
Example of point clouds from Synth4D using the simulated Velodyne (a) HDL32E and (b) HDL64E. 
}
\label{fig:dataset_examples}
\vspace{-.4cm}
\end{figure}

To enable full compatibility with SemanticKITTI \cite{behley2019iccv} and nuScenes \cite{nuscenes2019}, we present a new synthetic dataset, namely Synth4D, which we created using the CARLA simulator \cite{Dosovitskiy17}. 
Tab.~\ref{tab:dataset_comparison} compares Synth4D to the other synthetic datasets.
Synth4D is composed of two sets of point cloud sequences, one compatible with SemanticKITTI and one compatible with nuScenes.
Each set is composed of 20K labelled point clouds.
Synth4D is captured using a vehicle navigating in four scenarios, \textit{i.e.}, town, highway, rural area and city.
Because UDA requires consistent labels between source and target, we mapped the labels of Synth4D with those of SemanticKITTI/nuScenes using the original instructions given to annotators \cite{behley2019iccv, nuscenes2019}, thus producing eight macro classes: \textit{vehicle}, \textit{pedestrian}, \textit{road}, \textit{sidewalk}, \textit{terrain}, \textit{manmade}, \textit{vegetation} and \textit{unlabelled}.
Fig.~\ref{fig:dataset_examples} shows examples of annotated point clouds from Synth4D.
See Supp. Mat. for more details.



\section{SF-OUDA}
\vspace{-.2cm}
We formulate the problem of SF-OUDA for 3D point cloud segmentation as follows.
Given a deep network model $F_\mathcal{S}$ that is pre-trained with supervision on the source domain $\mathcal{S}$, we aim to adapt $F_\mathcal{S}$ on the target domain $\mathcal{T}$ given an unlabelled point cloud stream as input.
$F_\mathcal{S}$ is pre-trained using the source data $\Gamma_\mathcal{S} = \{ (X^i_\mathcal{S}, Y^i_\mathcal{S}) \}_{i=1}^{M_\mathcal{S}}$, where $X^i_\mathcal{S}$ is a synthetic point cloud, $Y^i_\mathcal{S}$ is the segmentation mask of $X^i_\mathcal{S}$ and $M_\mathcal{S}$ is the number of available synthetic point clouds.
Let $X^t_\mathcal{T}$ be a point cloud of our stream at time $t$ and $F^t_\mathcal{T}$ be the target model adapted using $X^t_\mathcal{T}$ and $X^{t-w}_\mathcal{T}$, with $w > 0$.
$Y_\mathcal{T}$ is the set of unknown target labels and $C$ is the number of classes contained in $Y_\mathcal{T}$.
The source classes and the target classes are coincident.

\subsection{Our approach}\label{sec:sfouda}
\begin{figure}[t]
    \centering
    \includegraphics[width=0.8\textwidth]{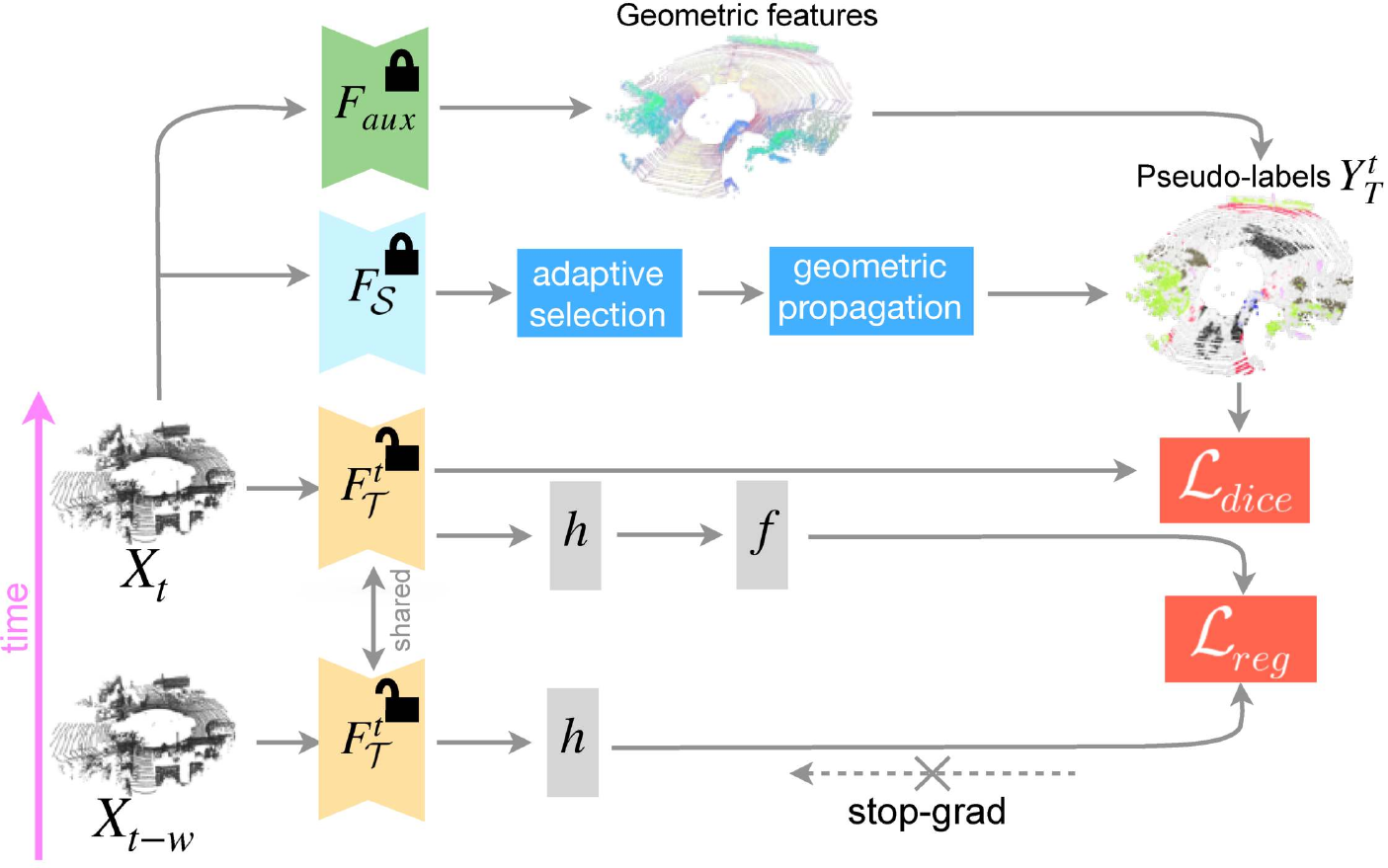}
    \vspace{-.4cm}
    \caption{Overview of \ourmethod. A source pre-trained model $F_\mathcal{S}$ selects \textit{seed pseudo-labels} through our adaptive-selection approach.
    An auxiliary model $F_{aux}$ extracts geometric features to guide pseudo-label propagation. 
    $\mathcal{L}_{dice}$ is minimised over the pseudo-labels $Y^t_T$. 
    In parallel, semantic smoothness is enforced with $\mathcal{L}_{reg}$ over time. 
    (\faLock) frozen parameters.
    (\reflectbox{\faLockOpen}) learnable parameters.}
    \label{fig:main_chart}
    \vspace{-.4cm}
\end{figure}

\vspace{-.2cm}
The input to \ourmethod is the point cloud $X^t_\mathcal{T}$ and an already processed point cloud $X^{t-w}_\mathcal{T}$.
These point clouds are used to adapt $F_\mathcal{S}$ to $\mathcal{T}$ through self-supervision (Fig.~\ref{fig:main_chart}).
The input is processed by two modules.
The first module aims to create labels for self-supervision by segmenting $X^t_\mathcal{T}$ with the source model $F_\mathcal{S}$.
Because these labels are produced by an unsupervised deep network, we refer to them as \emph{pseudo-labels}.
We select a subset of segmented points that have reliable pseudo-labels through an adaptive selection criteria, and propagate them to less reliable points.
The propagation uses geometric similarity in the feature space to increase the number of pseudo-labels available for self-supervision.
To this end, we use an auxiliary deep network ($F_{aux}$) that is specialized in extracting geometrically-informed representations from 3D points.
The second module aims to encourage temporal regularization of semantic information between $X^t_\mathcal{T}$ and $X^{t-w}_\mathcal{T}$.
Unlike recent works~\cite{huang2021spatio}, where a global point cloud descriptor of the scene is learnt, we exploit a self-supervised framework based on stop gradient \cite{chen2021exploring} to ensure smoothness over time.
Self-supervision through pseudo-label geometric propagation and temporal regularization are concurrently optimized to achieve the desired domain adaptation objective (Sec.~\ref{sec:final_loss}).

\noindent\textbf{Adaptive pseudo-label selection.}
An accurate selection of pseudo-labels is key to reliably adapt a model.
In dynamic real-world scenarios, where new structures appear/disappear in/from the LiDAR field of view, traditional pseudo-labeling techniques~\cite{chen2018domain, shin2020two} can suffer from unexpected variations of class distributions, producing overconfident incorrect pseudo-labels and making more populated classes prevail on others~\cite{zou2018unsupervised, zou2019confidence}.
We overcome this problem by designing a class-balanced adaptive-thresholding strategy to choose reliable pseudo-labels.
First, we compute an uncertainty index to filter out likely unreliable pseudo-labels.
Second, we apply a different threshold for each class based on the uncertainty index distribution.
This uncertainty index is directly related to the robustness of the output class distribution for each point.
Robust pseudo-labels can be extracted from those points that consistently provide similar output distributions under different dropout perturbations~\cite{kendall2015bayesian}.
We found that this approach works better than alternative confidence based approaches \cite{zou2018unsupervised, zou2019confidence}.

Given the point cloud $X^t_\mathcal{T}$, we perform $J$ iterations of inference with $F_\mathcal{S}$ by using dropout and obtain
\begin{equation}\label{eq:mean}
    p_\mathcal{T}^t = \frac{1}{J} \sum_{j=1}^J p \left( F_\mathcal{S} | X_\mathcal{T}^t , d_j \right),
\end{equation}
where $p_\mathcal{T}^t$ is the averaged output distribution of $F_\mathcal{S}$ given $X_\mathcal{T}^t$ and $d_j$, \emph{i.e.} the dropout at $j$-th iteration.
We compute the uncertainty index $\nu_\mathcal{T}^t$ as the variance over the $C$ classes of $p_\mathcal{T}^t$ as
\begin{equation}\label{eq:uncertainty_index}
    \nu_\mathcal{T}^t = E\left[\left(p_\mathcal{T}^t - \mu_\mathcal{T}^t \right)^2 \right],
\end{equation}
where $\mu_\mathcal{T}^t = E[p_\mathcal{T}^t]$ is the expected value of $p_\mathcal{T}^t$.
Then, we select the least uncertain points by using a different uncertainty threshold for each class.
Let $\lambda_c^t$ be the uncertainty threshold of class $c$ at time $t$.
Since $\nu_\mathcal{T}^t$ defines the uncertainty for each point, we group $\nu_\mathcal{T}^t$ values per class and compute $\lambda_c^t$ as the $a$-th percentile of $\nu_\mathcal{T}^t$ for class $c$.
Therefore, at time $t$ and for class $c$, we select only those pseudo-labels having the corresponding uncertainty index lower than $\lambda_c^t$ and use the corresponding pseudo-labels as \emph{seed pseudo-labels}.



\begin{figure}[t]
\centering
\begin{tabular}{@{}c@{}c}
    \begin{overpic}[width=.27\textwidth]{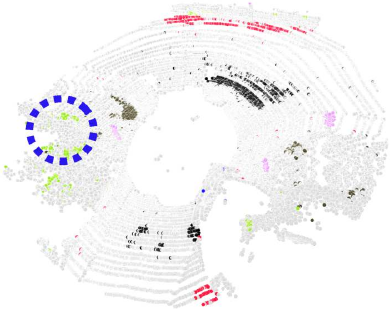}
    \put(10,-2){\color{black}\scriptsize\textbf{(a)}}
    \end{overpic}&
    \hspace{1.3cm}
    \begin{overpic}[width=.31\textwidth]{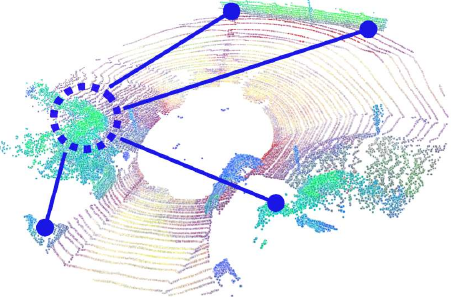}
    \put(10,-2){\color{black}\scriptsize\textbf{(b)}}
    \end{overpic}
\end{tabular}
\vspace{-.2cm}
\caption{Example of geometric propagation: a) starting from \textit{seed pseudo-labels}, b) geometric features are used to expand labels toward geometrically consistent regions.}
\label{fig:propagation}
\vspace{-.4cm}
\end{figure}

\vspace{.1cm}
\noindent\textbf{Geometric pseudo-label propagation.}
Typically, seed pseudo-labels are few and uninformative for the adaptation of the target model -- the deep network is already confident about them.
Therefore, we aim to propagate these pseudo-labels to potentially informative points.
This is challenging because the model may drift during adaptation.
We propose to use the features produced by an auxiliary geometrically-informed encoder $F_{aux}$ to propagate seed pseudo-labels to geometrically-similar points.
Geometric features can be extracted using deep networks that compute 3D local descriptors \cite{Gojcic2019, Ao2021, Poiesi2021}.
3D local descriptors are compact representations of local geometries with great generalization abilities across domains.
Our intuition is that, while the propagation in the metric space may propagate only in the spatial neighborhood of seed pseudo-labels, the use of geometric features would allow us to propagate to geometrically similar points, which can be distant from their seeds in the metric space (Fig.~\ref{fig:propagation}).

Given a seed pseudo-labeled point $\tilde{\mathbf{x}}^t \in X^t_\mathcal{T}$, we compute a set of geometric similarities as
\begin{equation}\label{eq:geometric_distance}
    \mathcal{G}_{\tilde{\mathbf{x}}}^t = \lVert F_{aux}(\tilde{\mathbf{x}}^t) - F_{aux}(X^t_\mathcal{T}) \rVert_2,
\end{equation}
where $||\cdot||_2$ is the $l_2$-norm and $\mathcal{G}_{\tilde{\mathbf{x}}}^t$ is the set that contains the similarity values between $\tilde{\mathbf{x}}^t$ and all the other points of $X^t_\mathcal{T}$ (except $\tilde{\mathbf{x}}^t$).
Then, we select the points that correspond to top $K$ values in $\mathcal{G}_{\tilde{\mathbf{x}}}^t$ and assign the pseudo-label of $\tilde{\mathbf{x}}^t$ to them.
Let $Y^t_\mathcal{T}$ be the final set of pseudo-labels that we use for fine-tuning our model.

\vspace{.1cm}
\noindent\textbf{Self-supervised temporal consistency loss.}
While the vehicle moves, the LiDAR sensor samples the environment from different viewpoints generating point clouds with different point distributions due to clutter and/or occlusions.
As points of consecutive point clouds can be simply matched over time by using the vehicle's odometry \cite{geiger2012we, nuscenes2019}, we can reasonably consider local variations of point distributions as local augmentations with the same semantic information.
As a result, we can exploit recent self-supervised techniques to enforce temporal smoothness of our semantic features.

We begin by computing the set of corresponding points between $X^{t-w}_\mathcal{T}$ and $X^t_\mathcal{T}$ by using the vehicle's odometry.
Let $T_{t-w \rightarrow t} \in \mathbb{R}^{4\times4}$ be the rigid transformation (from odometry) that maps $X^{t-w}_\mathcal{T}$ in the reference frame of $X^t_\mathcal{T}$.
We define the set of corresponding point $\Omega^{t, t-w}$ as
\begin{align}\label{eq:odometry_search}
    \Omega^{t, t-w} = & \left\{ \{ \mathbf{x}^{t} \in X^{t}_\mathcal{T}, \mathbf{x}^{t-w} \in X^{t-w}_\mathcal{T} \} : \right. \nonumber\\
    &  \mathbf{x}^{t} = \mathtt{NN} \left( T_{t-w \rightarrow t} \circ \mathbf{x}^{t-w}, X^t_\mathcal{T} \right), \nonumber\\
    & \left. \lVert \mathbf{x}^{t} - \mathbf{x}^{t-w} \rVert_2 < \tau \right\},
\end{align}
where $\mathtt{NN}(n,m)$ is the nearest-neighbour search given the set $m$ and the query $n$,
$\circ$ is the operator that applies $T_{t-w \rightarrow t}$ to a 3D point and $\tau$ is a distance threshold.

We adapt the self-supervised learning framework proposed in SimSiam~\cite{chen2021exploring} to semantically smooth point clouds over time.
We add an encoder network $h(\cdot)$ and a predictor head $f(\cdot)$ to the target model $F_\mathcal{T}$ and minimize the negative cosine similarity between consecutive semantic representations of corresponding points.
Let $z^t \triangleq h(x^t)$ be the encoder features over the target backbone for $x^t$ and let $q^t \triangleq f(h(x^t))$ be the respective predictor features.
We minimize the negative cosine similarity as
\begin{equation}
    \mathcal{D}_{t\rightarrow{t-w}}(q^t, z^{t-w}) = - \frac{q^t}{\left\Vert q^t\right\Vert_2} \cdot \frac{z^{t-w}}{\left\Vert z^{t-w}\right\Vert_2}
\label{eq:consistency}
\end{equation}
Time consistency is symmetric in the backward direction, hence we use the corresponding point of $x^t$ from $\Omega^{t, t-w}$ and define our self-supervised temporal consistency loss as
\begin{equation}\label{eq:temp_reg}
    \mathcal{L}_{reg} = \frac{1}{2} \mathcal{D}_{t\rightarrow{t-w}}(q^t, z^{t-w}) + \frac{1}{2} \mathcal{D}_{t-w\rightarrow{t}}(q^{t-w}, z^{t})
\end{equation}
where stop-grad is applied on $z^{t}$ and $z^{t-w}$.

\subsection{Online model update}\label{sec:final_loss}
\vspace{-.2cm}
Classes are typically highly unbalanced in each point cloud, \textit{e.g.}, a pedestrian class may be $1\%$ the number of points of the \textit{vegetation} class.
To this end, we use the soft Dice loss \cite{Jadon2020} as we found it works well when classes are unbalanced.
Let $\mathcal{L}_{dice}$ be our soft Dice loss that uses the pseudo-labels selected though Eq.~\ref{eq:geometric_distance} as supervision.
We define the overall adaptation objective as $\mathcal{L}_{tot} = \mathcal{L}_{dice} + \mathcal{L}_{reg}$, where $\mathcal{L}_{reg}$ is our regularization loss defined in Eq.~\ref{eq:temp_reg}.

\section{Experiments}\label{sec:experiments}

\vspace{-.2cm}
\subsection{Experimental setup}

\vspace{-.1cm}
\noindent\textbf{Source and target datasets.} We pre-train our source models on Synth4D and SynLiDAR~\cite{synlidar}, and validate our approach on the official validation sets of SemanticKITTI \cite{behley2019iccv} and nuScenes \cite{nuscenes2019} (target domains).
In SemanticKITTI, we use the sequence $08$ that is composed of 4071 point clouds at 10Hz.
In nuScenes, we use 150 sequences, each composed of 40 point clouds at 2Hz.

\noindent\textbf{Implementation details.} We use MinkowskiNet as deep network for point cloud segmentation \cite{choy20194d}.
We use ADAM: initial learning rate of $0.01$ with exponential decay, batch-size $16$ and weight decay $10^{-5}$.
{As auxiliary network $F_{aux}$, we use the PointNet-based architecture proposed in~\cite{Poiesi2021} trained on Synth4D that outputs a geometric features (descriptor) for a given 3D point.}
For online adaptation, we fix the learning rate to $10^{-3}$ and do not use schedulers as they would require prior knowledge about the stream length.
Because we adapt our model on each new incoming point cloud, we use batch-size equal to $1$.
We set $J$=5, $a$=1, $\tau$=0.3cm and use $0.5$ dropout probability.
We set $K$=10, $w$=5 on SemanticKITTI, and $K$=5, $w$=1 on nuScenes.
Parameters are the same in all the experiments.

\noindent\textbf{Evaluation protocol.}
We follow the traditional evaluation procedure for online learning methods \cite{cesa2004generalization, zhan2020online}, \textit{i.e.}, we evaluate the model performance on a new incoming frame using the model adapted up to the previous frame.
We compute the Intersection over Union (IoU) \cite{rahman2016optimizing} and report the average IoU (mIoU) improvement over the source (averaged over all the target sequences).
We also evaluate the online version of our source model by fine-tuning it with ground-truth labels for all the points in the scene (target).
{We also evaluate the target upper bound (target) of our method obtained from the online finetuning of our source models over labelled target point clouds.}

\begin{table}[t]
    \centering
    \tabcolsep 6pt
    \caption{Synth4D $\rightarrow$ SemanticKITTI online adaptation. 
    Source: pre-trained source model (lower bound).
    We report absolute mIoU for Source and mIoU relative to Source for the other methods.
    Key. SF: Source-Free. UDA: Unsupervised DA. O: Online.}
    \label{tab:synth4d2kitti}
    \vspace{-.2cm}
    \resizebox{1\textwidth}{!}{%
    \begin{tabular}{lccccccccccc}
        \toprule
        \textbf{Model} & 
        \textbf{SF} & \textbf{UDA} & \textbf{O} & \textbf{vehicle} & \textbf{pedestrian} & \textbf{road} & \textbf{sidewalk} & \textbf{terrain} & \textbf{manmade} & \textbf{vegetation} & \textbf{Avg} \\
        \midrule
        \CC{sourcecolor}Source & \CC{sourcecolor}  & \CC{sourcecolor} & \CC{sourcecolor} & \CC{sourcecolor} 63.90 & \CC{sourcecolor} 12.60 & \CC{sourcecolor} 38.10 & \CC{sourcecolor} 47.30 & \CC{sourcecolor} 20.20 & \CC{sourcecolor} 26.10 & \CC{sourcecolor} 43.30 & \CC{sourcecolor} 35.93\\
        Target & \ding{51} &  & \ding{51} & +16.84 & +5.49 & +8.48 & +34.44 & +51.92 & +45.68 & +39.09 & +28.85\\
        \midrule
        ADABN~\cite{li2016revisiting} & \ding{51} & \ding{51} & & -7.80 & -2.00 & -10.20 & -18.60 & -7.70 & +5.80 & -0.70 & -5.89 \\
        RayCast~\cite{langer2020domain} & & \ding{51} & & +3.80 & -2.60 & -3.10 & -0.50 & +7.30 & +4.50 & +0.20 & +1.37  \\
        \midrule
        ProDA$^*$ & \ding{51}  & \ding{51} & \ding{51} & -57.77 & -12.34 & -37.36 & -46.95 & -19.97 & -25.62 & -42.48 & -34.64 \\
        SHOT$^*$ & \ding{51} & \ding{51} & \ding{51} & -62.44 & -12.00 & -28.27 & -40.20 & -20.00 & -25.47 & -42.55 & -32.99 \\
        ONDA~\cite{mancini2018kitting} & \ding{51} & \ding{51} & \ding{51} & -13.60 & -1.70 & -10.60 & -20.00 & -7.10 & +3.90 & -5.10 & -7.74\\
        CBST$^*$ & \ding{51} & \ding{51} & \ding{51} & -0.13	& \textbf{+0.58} & -1.00 & -1.12 & +0.88 & +1.69 & +1.03 & +0.28\\
        TPLD$^*$ & \ding{51}  & \ding{51} & \ding{51} & +0.36 & +1.18 & -0.76 & -0.71 & +0.95 & +1.74 & +1.15 & +0.56 \\
        \vspace{-.3cm} & & & & & \\
        \midrule
        \vspace{-.3cm} & & & & & \\
        \ourmethod(Ours) & \ding{51} & \ding{51} & \ding{51} & \textbf{+13.12} & -0.54 & \textbf{+1.19} & \textbf{+2.45} & \textbf{+2.78} & \textbf{+5.64} & \textbf{+5.54} & \textbf{+4.31}\\
        \bottomrule
    \end{tabular}
    }
    \vspace{-0.5cm}
\end{table}

\subsection{Benchmarking existing methods for SF-OUDA}\label{sec:approaching}
\vspace{-.1cm}
Because our approach is the first that specifically tackles SF-OUDA in the context of 3D point cloud segmentation, we perform an in-depth analysis of the literature to identify previous adaptation methods that can be re-purposed for SF-OUDA.
Additionally, we experimentally evaluate their effectiveness on the considered datasets. 
We identify three categories of methods, as detailed below. 

\noindent\textbf{Batch normalization-based methods} 
perform domain adaptation by considering different statistics for source and target samples within Batch Normalization (BN) layers. Here, we consider ADABN~\cite{li2016revisiting} and ONDA~\cite{mancini2018kitting}. ADABN~\cite{li2016revisiting} is a source-free adaptation method which operates by updating the BN statistics assuming that all target data are available (offline adaptation). 
ONDA~\cite{mancini2018kitting} is the online version of ADABN~\cite{li2016revisiting}, where the target BN statistics are updated online based on the target data within a mini-batch. This can be regarded as a SF-OUDA method. However, these approaches are general-purpose methods and have not been previously evaluated for 3D point cloud segmentation.


\noindent\textbf{Prototype-based adaptation methods} use class centroids, \textit{i.e.} prototypes, to generate target pseudo-labels that can be transferred to other samples via clustering.
We implement SHOT~\cite{liang2020we} and ProDA~\cite{zhang2021prototypical}.
SHOT~\cite{liang2020we} exploits Information Maximization (IM) to promote cluster compactness during offline adaptation. 
We implement SHOT by adapting the pre-trained model with the proposed IM loss online on each incoming target point cloud. 
ProDA~\cite{zhang2021prototypical} adopts a centroid-based weighting strategy to denoise target pseudo-labels, while also considering supervision from source data. We adapt ProDA to SF-OUDA by applying the same weighting strategy but removing source data supervision. We update target centroids at each incremental learning step.
We refer to our SF-OUDA version of SHOT and PRODA as SHOT$^*$ and ProDA$^*$, respectively.
%


\noindent\textbf{Self-training-based methods} exploit source model predictions to adapt on the target domain by re-training the
model. We implement CBST~\cite{zou2018unsupervised} and TPLD~\cite{shin2020two}.
CBST~\cite{zou2018unsupervised} relies on a prediction confidence to select the most reliable pseudo labels. A confidence threshold is computed offline for each target class to avoid class unbalance. Our implementation of CBST, which we denote as CBST$^*$, uses the same class balance selection strategy but updates the thresholds online on each incoming frame. Moreover, no source data are considered as we are in a SF-OUDA setting. 
TPLD~\cite{shin2020two}, originally designed for 2D semantic segmentation, uses the pseudo-label selection mechanism in~\cite{zou2018unsupervised} but introduces a pixel pseudo label densification process. We implement TPLD by removing source supervision and replace the densification procedure with a 3D spatial nearest-neighbor propagation.
Our version of TPLD is denoted as TPLD$^*$.

Besides re-purposing existing approaches for SF-OUDA, we also evaluate an additional baseline, \textit{i.e.} the rendering-based method RayCast \cite{langer2020domain}. 
This approach is based on the idea that target-like data can be obtained with photorealistic rendering applied to the source point clouds. 
Thus, adaptation is performed by simply training on target-like data. 
While RayCast can be regarded as an offline adapation approach, we select it as it only requires the parameters of the real sensor to obtain target-like data from source point clouds.


\subsection{Results}

\begin{table*}[t]
    \centering
    \tabcolsep 6pt
    \caption{SynLiDAR $\rightarrow$ SemanticKITTI online adaptation. 
    Source: pre-trained source model (lower bound).
    We report absolute mIoU for Source and mIoU relative to Source for the other methods.
    Key. SF: Source-Free. UDA: Unsupervised DA. O: Online.}
    \label{tab:synlidar2kitti}
    \vspace{-.2cm}
    \resizebox{1.\textwidth}{!}{%
    \begin{tabular}{lccccccccccc}
        \toprule
        \textbf{Model} & \textbf{SF} & \textbf{UDA} & \textbf{O} & \textbf{vehicle} & \textbf{pedestrian} & \textbf{road} & \textbf{sidewalk} & \textbf{terrain} & \textbf{manmade} & \textbf{vegetation} & \textbf{Avg} \\
        \midrule
        \CC{sourcecolor}Source &\CC{sourcecolor}  & \CC{sourcecolor} &\CC{sourcecolor}  &\CC{sourcecolor} 59.80 &\CC{sourcecolor} 14.20 &\CC{sourcecolor} 34.90 &\CC{sourcecolor} 53.50 &\CC{sourcecolor} 31.00 &\CC{sourcecolor} 37.40 &\CC{sourcecolor} 50.50 & \CC{sourcecolor}40.19 \\
        Target & \ding{51} &  & \ding{51} & +21.32 & +8.09 & +11.51 & +28.13 & +40.46 & +33.67 & +30.63 & +24.83\\
        \midrule
        ADABN~\cite{li2016revisiting} & \ding{51} & \ding{51} & & +3.90 & -6.40 & -0.20 & -3.70 & -5.70 & +1.40 & +0.30 & -1.49\\
        RayCast~\cite{langer2020domain} & & \ding{51} & & - & - & - & - & - & - & - & -\\
        \midrule
        ProDA$^*$ & \ding{51}  & \ding{51} & \ding{51} & -53.30 & -13.79 & -33.83 & -52.78 & -30.52 & -36.68 & -49.29 & -38.60 \\
        SHOT$^*$ & \ding{51} & \ding{51} & \ding{51} & -57.83 & -12.64 & -24.80 & -46.02 & -30.80 & -36.83 & -49.32 & -36.89\\
        ONDA~\cite{mancini2018kitting} & \ding{51} & \ding{51} & \ding{51} & -2.90 & -6.40 & -2.20 & -8.80 & -7.60 & -1.20 & -6.70 & -5.11 \\
        CBST$^*$ & \ding{51} & \ding{51} & \ding{51} & +0.99 & -0.83 & +0.55 & +0.20 & +0.74 & -0.07 & +0.38 & +0.28 \\
        TPLD$^*$ & \ding{51}  & \ding{51} & \ding{51} & +0.90 & -0.48 & +0.59 & +0.33 & +0.84 & +0.07 & +0.37 & +0.37 \\
        \vspace{-.3cm} & & & & & \\
        \midrule
        \vspace{-.3cm} & & & & & \\
        \ourmethod(Ours) & \ding{51} &\ding{51} & \ding{51} & \textbf{+13.95} & -6.76 & \textbf{+3.26} & \textbf{+5.01} & \textbf{+3.00} & \textbf{+3.34} & \textbf{+4.08} & \textbf{+3.70}\\
        \bottomrule
    \end{tabular}
    }
\end{table*}

\begin{table*}[t]
    \centering
    \tabcolsep 6pt
    \caption{Synth4D $\rightarrow$ nuScenes online adaptation. 
    Source: pre-trained source model (lower bound).
    We report absolute mIoU for Source and mIoU relative to Source for the other methods.
    Key. SF: Source-Free. UDA: Unsupervised DA. O: Online.
    }
    \label{tab:synth4d2nuscenes}
    \vspace{-.2cm}
    \resizebox{1.0\textwidth}{!}{%
    \begin{tabular}{lccccccccccc}
        \toprule
        \textbf{Model} & 
        \textbf{SF} & \textbf{UDA} & \textbf{O} & \textbf{vehicle} & \textbf{pedestrian} & \textbf{road} & \textbf{sidewalk} & \textbf{terrain} & \textbf{manmade} & \textbf{vegetation} & \textbf{Avg} \\
        \midrule
        \CC{sourcecolor} Source & \CC{sourcecolor}  & \CC{sourcecolor}  & \CC{sourcecolor}  & \CC{sourcecolor} 22.54 & \CC{sourcecolor} 14.38 & \CC{sourcecolor} 42.03 & \CC{sourcecolor} 28.39 & \CC{sourcecolor} 15.58 & \CC{sourcecolor} 38.18 & \CC{sourcecolor} 54.14 & \CC{sourcecolor} 30.75\\
        Target & \ding{51} &  & \ding{51} & +3.76 & +0.92 & +9.41 & +16.95 & +19.79 & +10.92 & +10.71 & +10.35\\
        \midrule
        ADABN~\cite{li2016revisiting} & \ding{51} & \ding{51} & & +1.23 & -2.74 & -1.24 & +0.14 & +0.53 & +0.70 & +4.03 & +0.38\\
        RayCast~\cite{langer2020domain} & & \ding{51} & & -1.36 & -9.69 & -3.53 & -3.42 & -2.77 & -2.54 & -0.91 & -3.46\\
        \midrule
        ProDA$^*$ & \ding{51}  & \ding{51} & \ding{51} & +0.57 & \textbf{-1.40} & +0.73 & +0.09 & +0.71 & +0.40 & +0.91 & +0.29\\
        SHOT$^*$ & \ding{51} & \ding{51} & \ding{51} & \textbf{+0.82} & -1.77 & +0.68 & -0.05 & -0.70 & -0.54 & +1.09 & -0.07\\
        ONDA~\cite{mancini2018kitting} & \ding{51} & \ding{51} & \ding{51} & +0.34 & -1.90 & -1.19 & -0.62 & +0.18 & -0.40 & +0.58 & -0.43\\
        CBST$^*$ & \ding{51} & \ding{51} & \ding{51} & +0.37 & -2.61 & -1.35 & -0.79 & +0.19 & -0.36 & -0.45 & -0.71\\
        TPLD$^*$ & \ding{51}  & \ding{51} & \ding{51} & +0.65 & -1.90 & -0.96 & -0.39 & +0.43 & +0.07 & +0.86 & -0.18\\
        \vspace{-.3cm} & & & & & \\
        \midrule
        \vspace{-.3cm} & & & & & \\
        \ourmethod(Ours) & \ding{51} & \ding{51} & \ding{51} & +0.55 & -3.76 & \textbf{+1.64} & \textbf{+1.72} & \textbf{+2.28} & \textbf{+1.18} & \textbf{+2.36} & \textbf{+0.85}\\
        \bottomrule
    \end{tabular}
    }
    \vspace{-0.5cm}
\end{table*}

\vspace{-.1cm}
\noindent\textbf{Evaluating \ourmethod.} 
Tab.~\ref{tab:synth4d2kitti}, \ref{tab:synlidar2kitti} and \ref{tab:synth4d2nuscenes} report the results of our quantitative evaluation in the cases of Synth4D $\rightarrow$ SemanticKITTI, Synlidar $\rightarrow$ SemanticKITTI and Synth4D $\rightarrow$ nuScenes, respectively. The numbers in the tables indicate the improvement over the source model. 
\ourmethod achieves an average IoU improvement of +4.31 on Synth4D $\rightarrow$ SemanticKITTI, +3.70 on Synlidar $\rightarrow$ SemanticKITTI and +0.85 on Synth4D $\rightarrow$ nuScenes.
\ourmethod outperforms both offline and online methods by a large margin on Synth4D $\rightarrow$ SemanticKITTI and Synlidar $\rightarrow$ SemanticKITTI, while it achieves a lower improvement over Synth4D $\rightarrow$ nuScenes.
On SemanticKITTI, \ourmethod can effectively improve \emph{road}, \emph{sidewalk}, \emph{terrain}, \emph{manmade} and \emph{vegetation}.
\emph{vehicle} is the best performing class, which can achieve a mIoU above +13.
\emph{pedestrian} is the worst performing class on all the datasets.
\emph{pedestrian} is a challenging class because it is significantly unbalanced compared to the others, also in the source domain.
Although we attempted to mitigate the problem of unbalanced classes using adaptive thresholding and soft Dice loss, there are still situations that are difficult to address (see Sec.~\ref{conclusions} for details).
On nuScenes, the improvement is minor because at its lower resolutions makes patterns less distinguishable and more difficult to segment.

\noindent\textbf{Evaluating state-of-the-art methods.} We also analyze the performance of the existing methods discussed in Sec.~\ref{sec:approaching}. 
Batch-normalisation based methods perform poorly on all the datasets, with only ADABN~\cite{li2016revisiting} showing a minor improvement on nuScenes. 
We argue that non-i.i.d. batch samples arising in the online setting are playing an important role in this degradation, as they can have detrimental effects on models with BN layers~\cite{ioffe2017batch}.
SHOT$^*$ and ProDA$^*$ perform poorly in almost all the experiments, except on Synth4D $\rightarrow$ nuScenes where ProDA$^*$ achieves +0.29.
This minor improvement may be due to the short sequences of nuScenes (40 frames) making centroids less likely to drift. 
This does not occur in SemanticKITTI where the long sequence causes a rapid drift (see detailed in Sec.~\ref{sec:ablations}). 
CBST$^*$ and TPLD$^*$ improve on SemanticKITTI and perform poorly on nuScenes.
This can be ascribed to the noisy pseudo-labels that are selected using their confidence-based filtering approach.
Lastly, RayCast~\cite{langer2020domain} achieves +1.37 on Synth4D $\rightarrow$ SemanticKITTI, but underperform on Synth4D $\rightarrow$ nuScenes with a degradation of -3.46.
RayCast was originally proposed for real-to-real adaptation, therefore we believe that its performance may be affected by the large difference in point cloud resolution between Synth4D and nuScenes.
RayCast underperforms \ourmethod in the online setup, thus showing how offline solutions can fail in dynamic domains.
Note that RayCast cannot be evaluated using Synlidar, because Synlidar does not provide odometry information.

\subsection{In-depth analyses}\label{sec:ablations}
\vspace{-.2cm}
\noindent\textbf{Ablation study.}
Tab.~\ref{tab:component_ablation} shows the results of our ablation study on Synth4D $\rightarrow$ SemanticKITTI.
When we use only the adaptive pseudo-label selection (A) we can achieve +1.07 compared to the source.
When we combine A with the temporal regularization (T) we can further improve by +3.65.
Then we can achieve our best performance through the geometric propagation (P) of the pseudo labels.


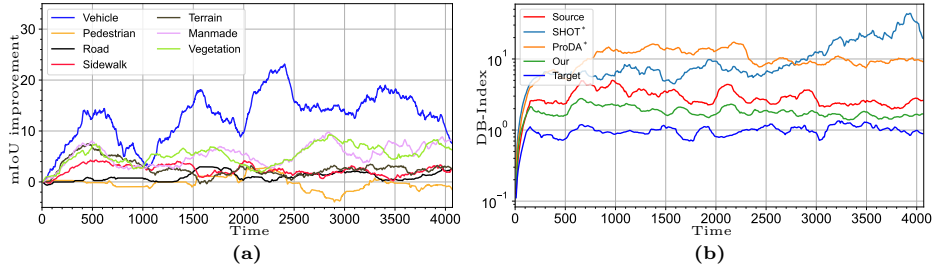
\begin{figure}[t]
\centering
    \begin{tabular}{cc}
    \raggedright
        \begin{overpic}[width=0.48\columnwidth]{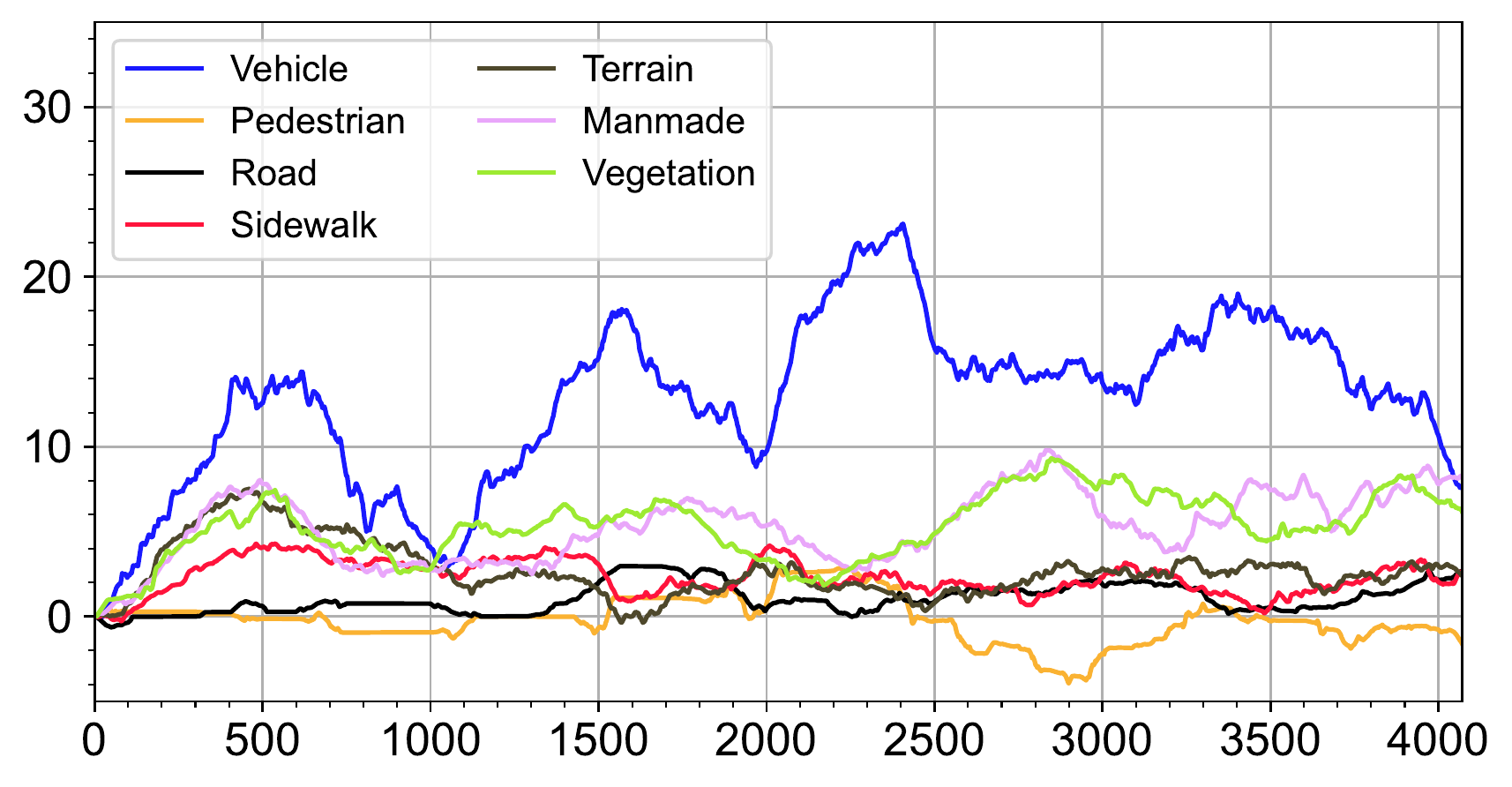}
        \put(-4, 15){\color{black}\tiny \rotatebox{90}{mIoU improvement}}
        \put(79, -4){\color{black}\tiny Time}
        \put(80, -13){\color{black}\scriptsize\textbf{(a)}}
        \end{overpic} & \hspace{0.1cm}
        \begin{overpic}[width=0.49\columnwidth]{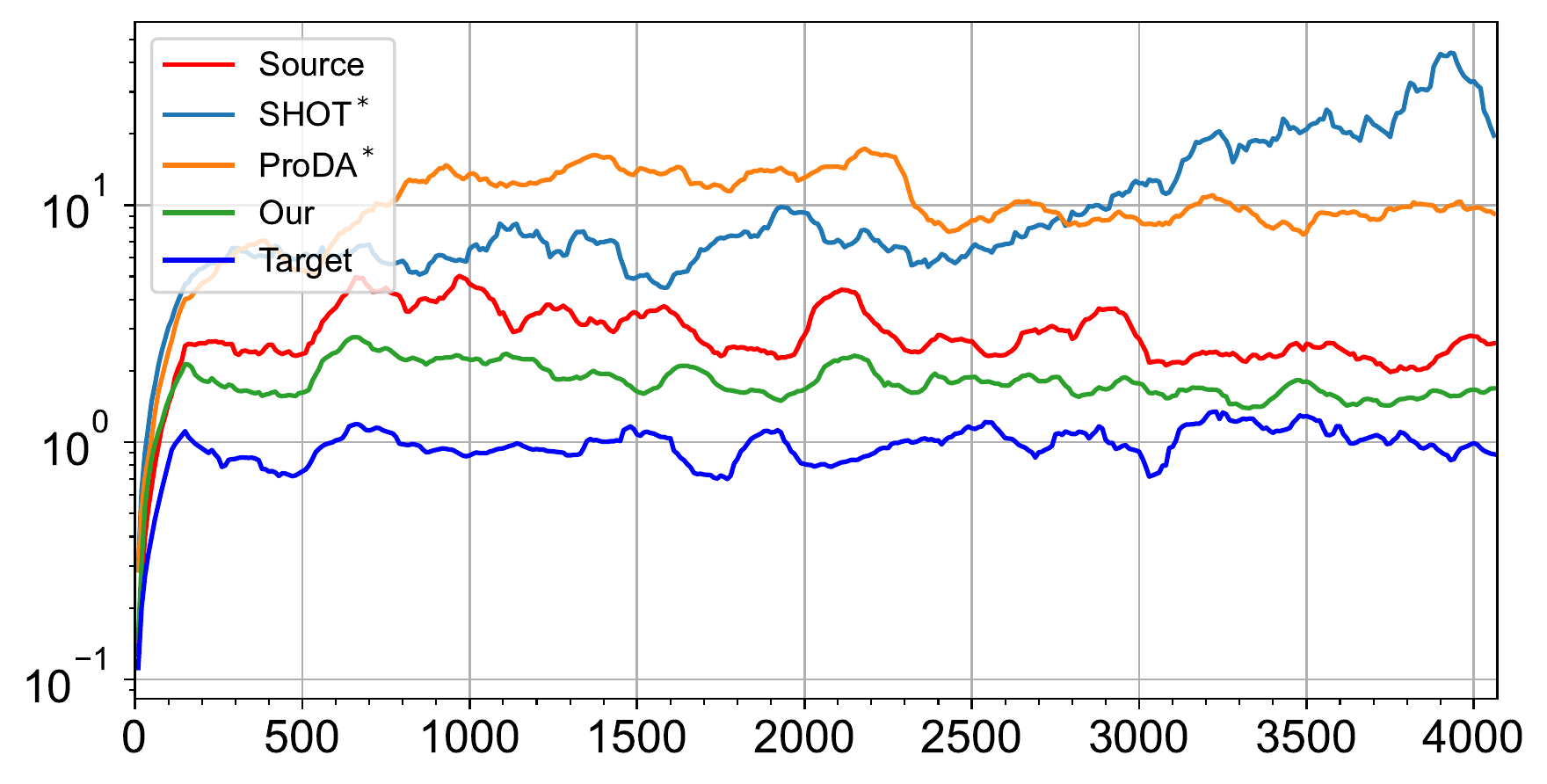}
        \put(-2, 30){\color{black}\tiny \rotatebox{90}{DB-Index}}
        \put(79, -4){\color{black}\tiny Time}
        \put(80, -13){\color{black}\scriptsize\textbf{(b)}}
        \end{overpic}
    \end{tabular}
    \vspace{.1cm}
    \caption{(a) Per-class improvement of \ourmethod over time on Synth4D$\rightarrow$SemanticKITTI. (b) DB-Index over time on Synth4D$\rightarrow$SemanticKITTI.
    The lower the DB-Index, the better the class separation of the features.}
    \label{fig:improvement_time_db_index}
\end{figure}

\noindent\textbf{Oracle study.}
We analyze the importance of using a reliable pseudo-label selection metric.
Tab.~\ref{tab:pseudo_accuracy} shows the pseudo-label accuracy as a function of the points that are selected as the $K$-th best candidates based on the distance from their centroids (as proposed in \cite{zhang2021prototypical}), confidence (as proposed in \cite{zou2018unsupervised}) and uncertainty (ours).
Centroid-based selection shows a low accuracy even at $K=1$, which tends to worsen as $K$ increases.
Confidence-based selection is more reliable than the centroid-based selection.
We found uncertainty-based selection to be more reliable at smaller values of $K$, which we deem to be more important than having more pseudo-labels but less reliable.

\begin{table}[t]
    \begin{minipage}{.49\textwidth}
    \tabcolsep 4pt
      \centering
        \caption{Synth4D$\rightarrow$SemanticKITTI ablation study of \ourmethod: (A) Adaptive thresholding; (A+T) A + Temporal consistency; (A+T+P) A+T + geometric Propagation.}
        \vspace{.05cm}
        \label{tab:component_ablation}
        \resizebox{.9\textwidth}{!}{
         \begin{tabular}{ccccc}
        \toprule
        \textbf{Source} & \textbf{Target} & \textbf{A} & \textbf{A+T} & \textbf{A+T+P} \\
        \midrule
        35.95 & +28.85 & +1.07 & +3.65 & +4.31 \\
        \bottomrule
    \end{tabular}
    }
    \end{minipage}%
    \hspace{.2cm}
    \begin{minipage}{.48\textwidth}
      \caption{Oracle study on Synth4D $\rightarrow$ SemanticKITTI that compares the accuracy of different pseudo-label selection metrics: Centroid, Confidence and Uncertainty.}
      \vspace{-.2cm}
      \label{tab:pseudo_accuracy}
      \centering
      \resizebox{.9\textwidth}{!}{%
        \begin{tabular}{lccc}
        \toprule
        & \textbf{Centroid} & \textbf{Confidence} & \textbf{Uncertainty} \\
        \midrule
        Top-1 & 38.1 & 66.7 & 76.1 \\
        Top-10 & 43.8 & 61.4 & 69.7 \\
        \bottomrule
        \end{tabular}
      }
    \end{minipage}
\end{table}


\noindent\textbf{Per-class temporal behavior.}
Fig.~\ref{fig:improvement_time_db_index}\textcolor{red}{a} shows the mIoU over time for each class on Synth4D $\rightarrow$ SemanticKITTI.
We can observe that six out of seven classes have a steady improvement:
\textit{vehicle} is the best performing class, followed by \textit{vegetation} and \textit{manmade}.
Drops in mIoU are typically due to sudden geometric variations of the point cloud, \textit{e.g.}, a road junction after a straight road, or a jammed road after a empty road.
\textit{pedestrian} confirms to be the most challenging class.

\noindent\textbf{Temporal compactness of features.}
We assess how well points are organized in the feature space over time.
We use the DB Index (DBI) that is typically used in clustering to measures the feature intra- and inter-class distances \cite{davies1979cluster}.
The lower the DBI, the better the quality of the features.
We use SHOT$^*$ and ProDA$^*$ as comparisons with our method, and the source and target models as references.
Fig.~\ref{fig:improvement_time_db_index}\textcolor{red}{b} shows the DBI variations over time.
SHOT$^*$ behavior is typical of a drift, as features of different classes become interwoven.
ProDA$^*$ does not drift, but it produces features that are worse than the source model.
Our approach is between source and target models, with a tendency to get closer to target.


\noindent\textbf{Different 3D local descriptors.}
We assess the effectiveness of different 3D local descriptors.
We test FPFH~\cite{rusu2009fast} (handcrafted) and FCGF~\cite{choy2019fully} (deep learning) descriptors.
\ourmethod achieves +3.56 mIoU with FPFH, +4.12 mIoU with FCGF and +4.31 mIoU with DIP.
This is inline with the experiments shown in~\cite{poiesi2022dip}, where DIP shows a superior generalization capability across domains than FCGF.

\noindent\textbf{Performance with global features.}
We assess the \ourmethod performance on Synth4D$\rightarrow$SemanticKITTI when the global temporal consistency loss proposed in STRL~\cite{huang2021spatio} is used instead of our per-point loss (Eq.~\ref{eq:consistency}). 
This variation achieves $+1.74$ mIoU, showing that per-point temporal consistency is key.

\begin{figure}[t]
\centering
    \begin{tabular}{cc}
    \raggedright
        \begin{overpic}[width=0.45\columnwidth]{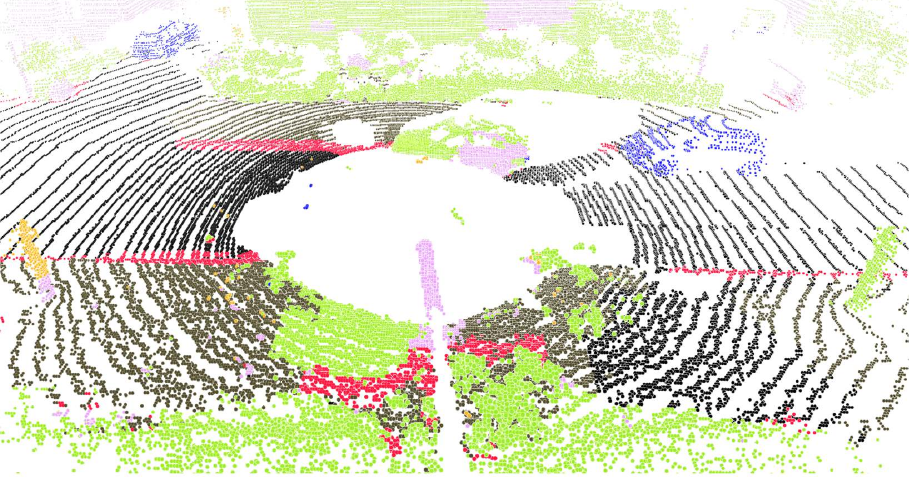}
        \put(65,85){\color{black}\footnotesize \textbf{source}}
        \put(-8,35){\color{black}\footnotesize \rotatebox{90}{\textbf{large}}}
        \end{overpic} &
        \begin{overpic}[width=0.45\columnwidth]{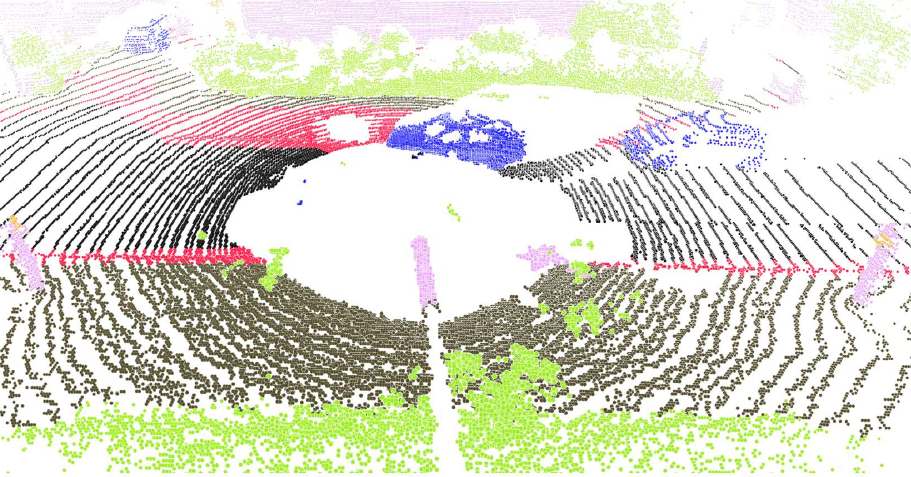}
        \put(65,85){\color{black}\footnotesize \textbf{ours} }
        \end{overpic}\\
        \begin{overpic}[width=0.45\columnwidth]{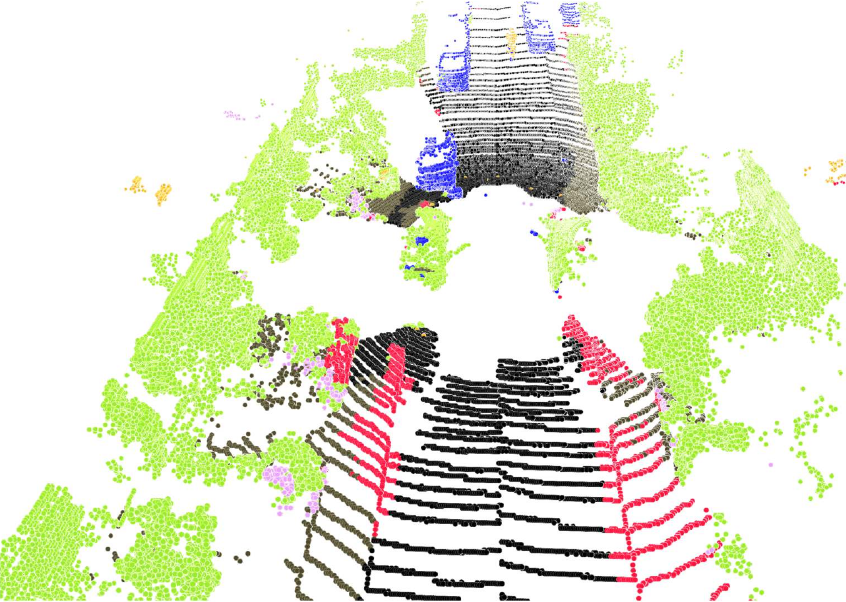}
        \put(-8,30){\color{black}\footnotesize \rotatebox{90}{\textbf{medium}}}
        \end{overpic} &  
        \begin{overpic}[width=0.45\columnwidth]{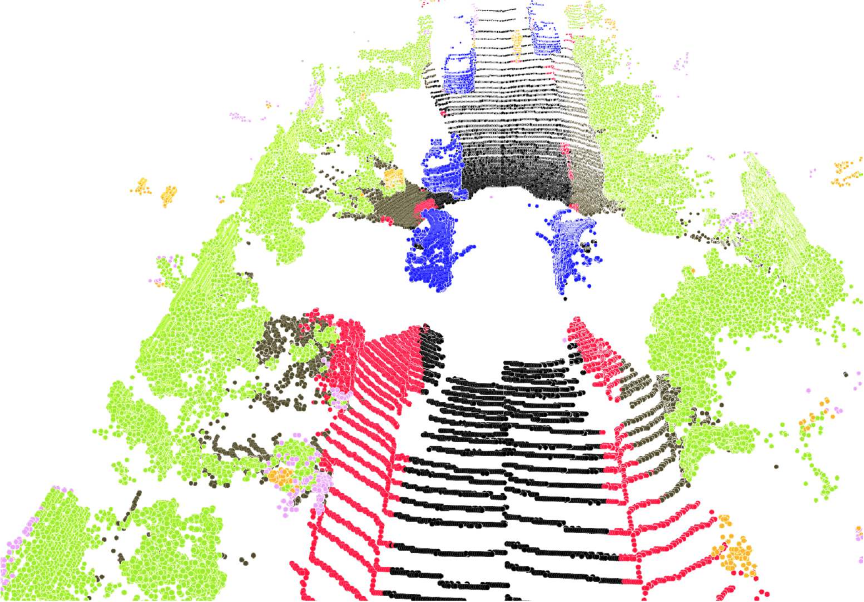}
        \end{overpic}\\
        \begin{overpic}[width=0.45\columnwidth]{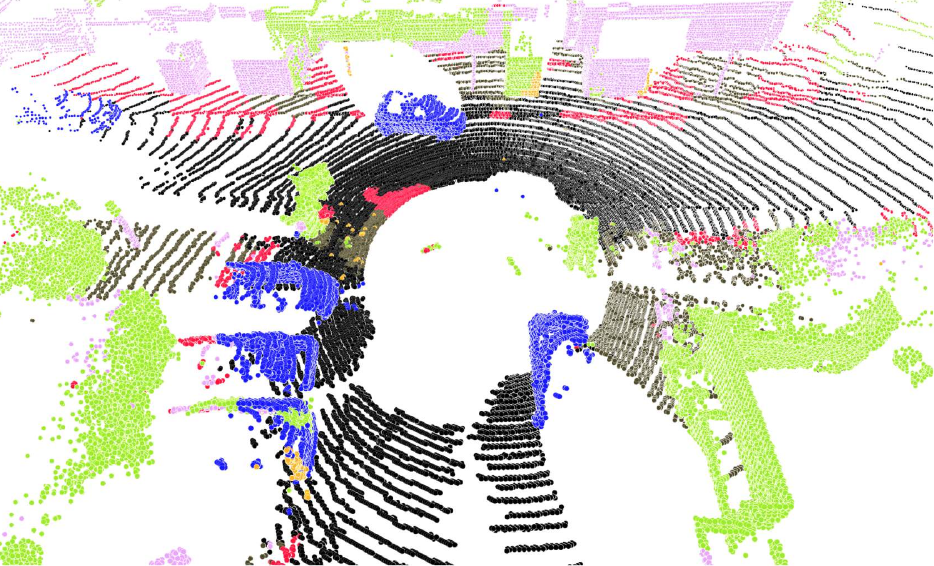}
        \put(-8,30){\color{black}\footnotesize \rotatebox{90}{\textbf{small}}}
        \end{overpic} &  
        \begin{overpic}[width=0.45\columnwidth]{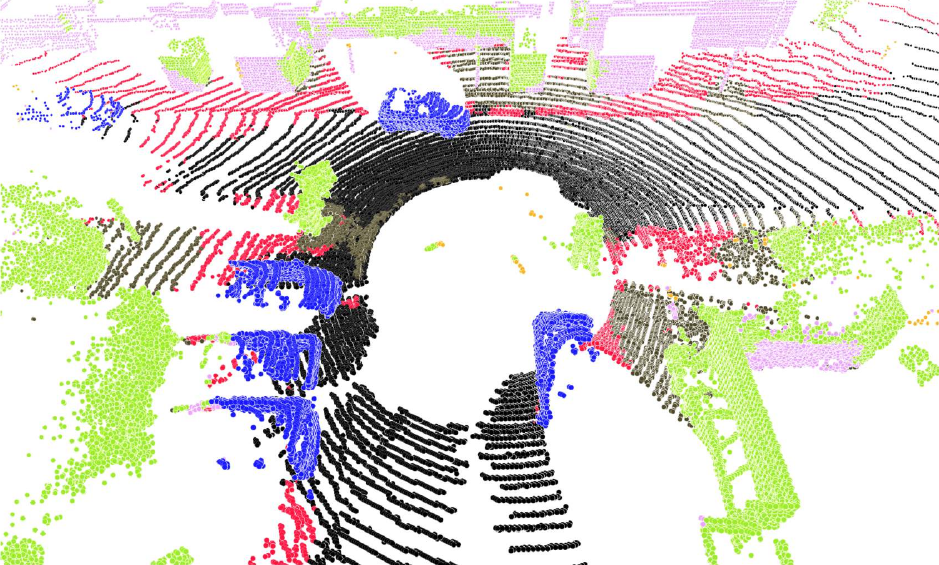}
        \end{overpic}
    \end{tabular}
    \begin{overpic}[width=1.\columnwidth]{images/dataset/legend_h.pdf}
    \end{overpic}
    \vspace{-.7cm}
    \caption{Results on Synth4D$\rightarrow$SemanticKITTI with three different ranges of mIoU improvements, i.e., large (+27.2), medium (+10.0) and small (+5.1).}
    \label{fig:qualitative}
    \vspace{-0.5cm}
\end{figure}

\noindent\textbf{Qualitative results.}
Fig.~\ref{fig:qualitative} shows the comparison between \ourmethod and the source model on on Synth4D$\rightarrow$SemanticKITTI.  
The first row shows frame $178$ of SemanticKITTI with an improvement of $+27.14$ mIoU (large).
The classes \textit{vehicle}, \textit{sidewalk} and \textit{terrain} are incorrectly segmented by the source model, we can see a significant improvement in segmentation on these classes after adaptation.
The second and third rows show frame $1193$ and frame $2625$ with an improvement of $+10.00$ mIoU (medium) and $+4.99$ mIoU (small).
Improvements are visible after adaptation in the classes \textit{vehicle}, \textit{sidewalk} and \textit{road}.
The last row shows a segmentation drift for \textit{road} that is caused by incorrect pseudo-labels.

\section{Discussions}\label{conclusions}
\vspace{-.2cm}
\noindent\textbf{Conclusions}.
We studied for the first time the problem of SF-OUDA for 3D point cloud segmentation in a synthetic-to-real setting. 
We experimentally showed that existing approaches do not suffice in coping with domain shift in this scenario. 
We presented \ourmethod that relies on adaptive self-training and geometric-features propagation to address SF-OUDA.
We also introduced a novel synthetic dataset, namely Synth4D composed of two splits and matching the sensor setup of SemanticKITTI and nuScenes, respectively. 
Experiments on three different benchmarks showed that \ourmethod outperforms state-of-the-art approaches.

\noindent\textbf{Limitations.} 
\ourmethod limitations are related to geometric propagation and long-tailed classes.
If objects of different classes share similar geometric structures, the geometric propagation may be deleterious.
This can be mitigated by using another sensor modality (e.g.~RGB) or by accounting for multi-scale signals to exploit context information.
If severe class unbalance occurs, semantic segmentation accuracy may be affected, e.g.~\textit{pedestrian} class in Tabs.~\ref{tab:synth4d2kitti}-\ref{tab:synth4d2nuscenes}.
This can be mitigated by re-weighting the loss through a class-balanced term (computed on the source).

\vspace{.2cm}
\noindent\textbf{Acknowledgments.}
This work was partially supported by OSRAM GmbH, by the Italian Ministry of Education, Universities and Research (MIUR) ``Dipartimenti di Eccellenza 2018-2022'', by the EU JPI/CH SHIELD project, by the PRIN project PREVUE (Prot. 2017N2RK7K), the EU ISFP PROTECTOR (101034216) project and the EU H2020 MARVEL (957337) project and, it was carried out in the Vision and Learning joint laboratory of FBK and UNITN.


\clearpage
%
%
\bibliographystyle{splncs04}
\bibliography{egbib}
\end{document}


\pagestyle{headings}
\mainmatter
\def\ECCVSubNumber{2016}  

\title{Supplementary Material\\
GIPSO: 
Geometrically Informed Propagation for 
Online Adaptation in 3D LiDAR Segmentation} 


\titlerunning{Abbreviated paper title}
%
\titlerunning{GIPSO: Geometrically Informed Prop. for Online Adapt. in 3D LiDAR Seg.}
%

\author{Cristiano Saltori\inst{1} \and
Evgeny Krivosheev \inst{1} \and
Stéphane Lathuilière\inst{2} \and
Nicu Sebe\inst{1} \and
Fabio Galasso\inst{3} \and
Giuseppe Fiameni\inst{4} \and
Elisa Ricci\inst{1,5} \and
Fabio Poiesi\inst{5}
}

\authorrunning{C. Saltori et al.}
%
\institute{University of Trento, Trento, Italy \and
LTCI, Télécom-Paris, Intitute Polytechnique de Paris, Palaiseau, France \and
Sapienza University of Rome, Rome, Italy \and
NVIDIA AI Technology Center \and
Fondazione Bruno Kessler, Trento, Italy\\
\email{cristiano.saltori@unitn.it}}

\maketitle

\section{Introduction}
We provide supplementary material in support of the main paper. The content is organized as follows:
\begin{itemize}
    \item Sec.~\ref{sec:architecture} reports the architecture details of the main modules used in GIPSO;
    \item Sec.~\ref{sec:gipso} provides additional ablations of \ourmethod, analysing the performance with a different propagation size and time-window length;
    \item Sec.~\ref{sec:improving} goes beyond \ourmethod and shows that our proposed strategies can be used to improve baselines in SF-OUDA;
    \item Sec.~\ref{sec:mapping} reports the class mapping used in our experiments for compatibility between synthetic and real domains;
    \item In Sec.~\ref{sec:qualitative}, additional qualitative results are reported on Synth4D $\rightarrow$ SemanticKITTI, SynLiDAR $\rightarrow$ SemanticKITTI, and Synth4D $\rightarrow$ nuScenes.
    
\end{itemize}

\section{Architecture details} \label{sec:architecture}
We implemented \ourmethod in PyTorch by using minkowski/sparse convolutions in MinkowskiEngine~\cite{choy20194d}. For the backbone and segmentation network we used the existing implementation of MinkUNet18~\cite{choy20194d} by setting the dimension of the input space to $D=3$, \textit{i.e.} the dimensionality of an input point cloud. 
For the self-supervised temporal consistency loss (Sec.~{\color{red}4}, Eq.~6)
we implemented the encoder $h()$ with two consecutive MinkowskiConvolution layers interleaved by a ReLU activation function and a batch-normalization layer. The input size of the first layer is set to $96$ - the output feature size of the backbone network - while the output size is set to $128$. The last encoding layer is set to have the same input and output size of $128$.
We implemented the predictor $f()$ with the same structure of $h()$ with the difference that input and output sizes are set to $128$. In both $h()$ and $f()$ we used a kernel of size $1$, biases activated and $D=3$.

\section{\ourmethod components}\label{sec:gipso}
We provide two additional ablation studies to complement the ablation study in the main manuscript in Sec.~{\color{red}5.4}.
We perform an ablation study for different components of \ourmethod on Synth4D $\rightarrow$ SemanticKITTI.
Sec.~\ref{sec:propagation} reports the results when the propagation size $K$ is increased up to $100$ for each seed pseudo-label. 
Sec.~\ref{sec:time} reports how \ourmethod performs by varying the time window $w$. 
Results report the performance on Source (gray) in absolute mIoU while the others are reported as relative mIoU improvement over the Source model. Target is the supervised upper bound of our task in our setting.

\subsection{Propagation size}\label{sec:propagation}
We study the effect of different propagation steps by using our geometry-based propagation. Tab.~\ref{tab:propagation_size} shows the results with a $K$ of $1,5,10,50,100$. 
We can see that mIoU starts to decrease when a higher number of propagation steps are used, i.e., $K=50$, whereas we reach the best improvement of $+4.31$ with $K=10$. 
These results show that $K$ should be set such that to both preserve pseudo-labelling accuracy while propagating seed labels towards new informative points.

\begin{table*}[t]
    \centering
    \tabcolsep 6pt
    \caption{Online adaptation on Synth4D $\rightarrow$ SemanticKITTI with different propagation size $K$.
    }
    \label{tab:propagation_size}
    \vspace{-.3cm}
    \resizebox{\textwidth}{!}{%
    \begin{tabular}{lccccccccc}
        \toprule
        \textbf{Model} & \textbf{$K$} & \textbf{vehicle} & \textbf{pedestrian} & \textbf{road} & \textbf{sidewalk} & \textbf{terrain} & \textbf{manmade} & \textbf{vegetation} & \textbf{Avg} \\
        \midrule
        \CC{sourcecolor} Source & \CC{sourcecolor}- & \CC{sourcecolor} 22.54 & \CC{sourcecolor} 14.38 & \CC{sourcecolor} 42.03 & \CC{sourcecolor} 28.39 & \CC{sourcecolor} 15.58 & \CC{sourcecolor} 38.18 & \CC{sourcecolor} 54.14 & \CC{sourcecolor} 30.75\\
        Target & - &+3.76 & +0.92 & +9.41 & +16.95 & +19.79 & +10.92 & +10.71 & +10.35\\
        \midrule
        Ours & 1 & +14.18 & -1.13 & +1.08 & +2.11 & +2.74 & +5.49 & +5.39 & +4.27\\
        Ours & 5 & +13.42 & -0.51 & +0.91 & +2.16 & +2.66 & +5.54 & +5.62 & +4.26\\
        Ours & 10 & +13.12 & -0.54 & +1.19 & +2.45 & +2.78 & +5.64 & +5.54 & +4.31\\
        Ours & 50 & +12.01 & -1.00 & +0.73 & +2.01 & +3.02 & +5.51 & +5.66 & +3.99\\
        Ours & 100 & +12.25 & -2.49 & +0.62 & +1.93 & +3.39 & +5.99 & +5.68 & +3.91\\
        \bottomrule
    \end{tabular}
    }
\end{table*}

\subsection{Time-window length}\label{sec:time}
We study the effect of different time window length $w$ in our self-supervised temporal consistency loss. 
Tab.~\ref{tab:time_window} shows that $w$ should be selected neither too large ($w=8$) nor too small ($w=1$) for the best performance. 
The time window $w$ should be set based on the sampling rate of the sensor and the overlap between adjacent frames.

\begin{table*}[t]
    \centering
    \tabcolsep 6pt
    \caption{Online adaptation on Synth4D $\rightarrow$ SemanticKITTI with a different time window $w$.}
    \label{tab:time_window}
    \vspace{-.3cm}
    \resizebox{\textwidth}{!}{%
    \begin{tabular}{lccccccccc}
        \toprule
        \textbf{Model} & \textbf{$w$} & \textbf{vehicle} & \textbf{pedestrian} & \textbf{road} & \textbf{sidewalk} & \textbf{terrain} & \textbf{manmade} & \textbf{vegetation} & \textbf{Avg} \\
        \midrule
        \CC{sourcecolor} Source & \CC{sourcecolor}- & \CC{sourcecolor} 22.54 & \CC{sourcecolor} 14.38 & \CC{sourcecolor} 42.03 & \CC{sourcecolor} 28.39 & \CC{sourcecolor} 15.58 & \CC{sourcecolor} 38.18 & \CC{sourcecolor} 54.14 & \CC{sourcecolor} 30.75\\
        Target & - &+3.76 & +0.92 & +9.41 & +16.95 & +19.79 & +10.92 & +10.71 & +10.35\\
        \midrule
        Our & 1 & +9.73 & -0.63 & +0.56 & +1.79 & +2.86 & +4.88 & +4.27 & +3.35\\
        Our & 2 & +11.76 & -1.09 & +0.78 & +1.97 & +2.50 & +5.01 & +5.23 & +3.74\\
        Our & 3 & +12.89 & -0.37 & +0.79 & +1.84 & +2.70 & +5.20 & +5.12 & +4.02\\
        Our & 4 & +13.84 & -0.84 & +0.94 & +2.24 & +2.57 & +5.37 & +5.49 & +4.23\\
        Our & 5 & +13.12 & -0.54 & +1.19 & +2.45 & +2.78 & +5.64 & +5.54 & +4.31\\
        Our & 6 & +13.95 & -0.48 & +0.95 & +2.01 & +2.77 & +5.69 & +5.93 & +4.40\\
        Our & 7 & +13.32 & -0.90 & +1.11 & +2.16 & +3.14 & +5.43 & +5.74 & +4.28\\
        Our & 8 & +13.16 & -1.16 & +0.95 & +1.88 & +2.67 & +5.75 & +6.20 & +4.21\\
        \bottomrule
    \end{tabular}
    }
\end{table*}

\section{Improving state-of-the-art with \ourmethod}\label{sec:improving}
We show that our proposed modules also improve state-of-the-art methods, such as CBST~\cite{zou2018unsupervised}, ProDA~\cite{zhang2021prototypical} and, TPLD~\cite{zhang2021prototypical}, providing additional evidence that our propositions are steps forward in SF-OUDA not just in \ourmethod. 
First, we show that our adaptive sampling strategy can be used in state-of-the-art methods to obtain more reliable pseudo-labels. Second, we propose modifications to further improve baselines performance in SF-OUDA.
We propose the following modifications:
\begin{itemize}
    \item CBST$^*$ uses a confidence based sampling strategy to select class-balanced pseudo-labels. We improve CBST$^*$ by using our adaptive selection strategy based on uncertainty;
    \item TPLD$^*$ builds upon CBST$^*$ by increasing pseudo-label number through densification and voting. We improve TPLD$^*$ with our more robust adaptive pseudo-label selection and substitute the spatial nearest neighbor with our geometrically informed propagation strategy.
    \item ProDA$^*$ exploits a centroid-based weighting strategy to denoise pseudo-labels. Moreover, momentum update is performed between source $F_\mathcal{S}$ and target model $F_\mathcal{T}$. We improve ProDA$^*$ in its three main parts. First, we remove source model momentum update as it promotes domain drift. Second, we substitute pseudo-labelling with our iterative dropout based pseudo-labeling strategy. Third, we compute more robust centroids by considering the mean of point-features in our iterative pseudo-labelling strategy.
\end{itemize}

Tab.~\ref{tab:improving} shows that \ourmethod components can be used to successfully improve the performance of existing methods. 
ProDA$^*$ improves from $-32.63$ to $+1.48$, we deem this is due to the more robust centroid computation and to the lower adaptation drift obtained with a non-updated source model.
CBST$^*$ benefits from a better pseudo-label selection improving from $+0.28$ to $+1.07$. 
TPLD$^*$ benefits from a better pseudo-labels and the geometrically informed propagation improving from $*0.56$ to $+1.38$.

\begin{table*}[t]
    \centering
    \tabcolsep 6pt
    \caption{Ablation study on Synth4D $\rightarrow$ SemanticKITTI reporting the improvement of state-of-the-art methods by using \ourmethod adaptive selection strategy and propagation strategy.}
    \label{tab:improving}
    \vspace{-.2cm}
    \resizebox{\textwidth}{!}{%
    \begin{tabular}{lcccccccccc}
        \toprule
        \textbf{Model} & \textbf{vehicle} & \textbf{pedestrian} & \textbf{road} & \textbf{sidewalk} & \textbf{terrain} & \textbf{manmade} & \textbf{vegetation} & \textbf{Avg} \\
        \midrule
        \CC{sourcecolor} Source & \CC{sourcecolor} 22.54 & \CC{sourcecolor} 14.38 & \CC{sourcecolor} 42.03 & \CC{sourcecolor} 28.39 & \CC{sourcecolor} 15.58 & \CC{sourcecolor} 38.18 & \CC{sourcecolor} 54.14 & \CC{sourcecolor} 30.75\\
        Target & +3.76 & +0.92 & +9.41 & +16.95 & +19.79 & +10.92 & +10.71 & +10.35\\
        \midrule
        ProDA$^*$ & -58.92 & -12.08 & -36.74 & -45.32 & -15.46 & -20.69 & -39.24 & -32.63\\
        CBST$^*$ & -0.13 & 0.58 & -1.00 & -1.12 & 0.88 & 1.69 & 1.03 & 0.28 \\
        TPLD$^*$ & 0.36 & 1.18 & -0.76 & -0.71 & 0.95 & 1.74 & 1.15 & 0.56\\
        \midrule
        ProDA$^*$ (Ours)& 2.04 & 4.40 & 0.24 & 0.62 & 0.29 & 1.07 & 1.71 & 1.48\\
        CBST$^*$ (Ours)& 2.72 & -2.53 & -0.19 & 0.56 & 1.48 & 3.02 & 2.46 & 1.07\\
        TPLD$^*$ (Ours)& 2.81 & -2.33 & -0.05 & 0.65 & 2.30 & 3.44 & 2.82 & 1.38\\
        \bottomrule
    \end{tabular}
    }
\end{table*}

\section{Class mapping}\label{sec:mapping}
In Sec.~\ref{sec:synth4d} we detail the class mapping to make Synth4D compatible with SemanticKITTI \cite{behley2019iccv} and nuScenes \cite{nuscenes2019}. 
In Sec.~\ref{sec:synlidar} we report the class mapping used in SynLiDAR \cite{synlidar}.

\subsection{Synth4D}\label{sec:synth4d}
Tab.~\ref{tab:carla2synth4d} reports the class mapping from Cityscapes \cite{cordts2016cityscapes} format of CARLA \cite{Dosovitskiy17} to the classes of Synth4D.
Tab.~\ref{tab:kitti2synth4d} reports the class mapping from SemanticKITTI to Synth4D. Tab.~\ref{tab:nusc2synth4d} reports the class mapping from nuScenes to Synth4D.

Tab.~\ref{tab:carla2synth4d}-\ref{tab:nusc2synth4d} maps input labels into the eight Synth4D labels: \emph{vehicle, pedestrian, road, sidewalk, terrain, manmade, vegetation} and, \emph{unlabelled}. This class mapping corresponds to the label intersections between CARLA, SemanticKITTI and nuScenes. All the classes that do not intersect with other datasets are considered as \emph{unlabelled}.

\begin{table*}[t]
    \centering
    \caption{Class mapping from CARLA \cite{Dosovitskiy17} format to Synth4D.}
    \vspace{-.2cm}
    \begin{tabular}{cccc}
        \toprule
        CARLA-ID & CARLA-Name &  Synth4D-Name & Synth4D-ID\\
        \midrule
        0 & unlabelled & unlabelled & 0 \\
        1 & building & manmade & 6 \\
        2 & fences & manmade & 6 \\
        3 & other & unlabelled & 0 \\
        4 & pedestrian & pedestrian & 2 \\
        5 & pole & manmade & 6 \\
        6 & roadlines & road & 3 \\
        7 & road & road & 3 \\
        8 & sidewalk & sidewalk & 4\\
        9 & vegetation & vegetation & 7 \\
        10 & vehicle & vehicle & 1\\
        11 & wall & manmade & 6 \\
        12 & trafficsign & manmade & 6 \\
        13 & sky & unlabelled & 0 \\
        14 & ground & unlabelled & 0 \\
        15 & bridge & manmade & 6 \\
        16 & railtrack & manmade & 6 \\
        17 & guardrail & manmade & 6 \\
        18 & trafficlight & unlabelled & 0 \\
        19 & static & unlabelled & 0 \\
        20 & dynamic & unlabelled & 0 \\
        21 & water & unlabelled & 0 \\
        22 & terrain & terrain & 5 \\
        \bottomrule

    \end{tabular}
    \label{tab:carla2synth4d}
\end{table*}

\begin{table*}[t]
    \centering
    \caption{Class mapping from SemanticKITTI \cite{behley2019iccv} format to Synth4D.}
    \vspace{-.2cm}
    \begin{tabular}{cccc}
        \toprule
        SemanticKITTI-ID & SemanticKITTI-Name &  Synth4D-Name & Synth4D-ID\\
        \midrule
        0 & unlabelled & unlabelled & 0 \\
        1 &	car & vehicle & 1 \\
        2 &	bicycle & unlabelled & 0 \\
        3 &	motorcycle & unlabelled & 0 \\
        4 &	truck &	unlabelled & 0 \\
        5 &	other-vehicle &	unlabelled & 0 \\
        6 &	person & pedestrian & 2 \\
        7 &	bicyclist &	unlabelled & 0 \\
        8 &	motorcyclist & unlabelled & 0 \\
        9 &	road & road & 3 \\
        10 & parking& road & 3 \\
        11 & sidewalk & sidewalk & 4 \\
        12 & other-ground & unlabelled & 0 \\
        13 & building& manmade & 6 \\
        14 & fence & manmade & 6 \\
        15 & vegetation& vegetation & 7 \\
        16 & trunk & vegetation & 7 \\
        17 & terrain & terrain & 5 \\
        18 & pole &	manmade & 6 \\
        19 & traffic-sign & manmade & 6 \\
        \bottomrule
    \end{tabular}
    \label{tab:kitti2synth4d}
\end{table*}

\begin{table*}[t]
    \centering
    \caption{Class mapping from nuScenes \cite{nuscenes2019} format to Synth4D.}
    \vspace{-.2cm}
    \begin{tabular}{cccc}
        \toprule
        nuScenes-ID & nuScenes-Name &  Synth4D-Name & Synth4D-ID\\
        \midrule
        0 & unlabelled & unlabelled & 0 \\
        1 & barrier & unlabelled & 0 \\
        2 & bicycle & unlabelled & 0 \\
        3 & bus & unlabelled & 0 \\
        4 & car & vehicle & 1\\
        5 & construction-vehicle & unlabelled & 0 \\
        6 & motorcycle & unlabelled & 0\\
        7 & pedestrian & pedestrian & 2 \\
        8 & traffic-cone & unlabelled & 0 \\
        9 & trailer & unlabelled & 0 \\
        10 & truck & unlabelled & 0\\
        11 & driveable-surface & road & 3 \\
        12 & other-flat & unlabelled & 0 \\
        13 & sidewalk & sidewalk & 4\\
        14 & terrain & terrain & 5\\
        15 & manmade & manmade & 6 \\
        16 & vegetation & vegetation & 7\\
        \bottomrule
    \end{tabular}
    \label{tab:nusc2synth4d}
\end{table*}

Using the mapping in Tab.~\ref{tab:carla2synth4d}, the resulting class distributions for Synth4D are reported in Tab.~\ref{tab:dataset_stats}. It is important to notice that class distributions differ among sensors as they have been acquired with independent runs. During each run, the simulator is set to randomly initialise the ego-vehicle re-spawn position, agents' positions (i.e., vehicles and pedestrians) and agents' trajectories. Therefore, the same class distribution cannot be ensured.
\begin{table*}[t]
    \centering
    \tabcolsep 2pt
    \caption{Number of annotated points for each adaptation category for the simulated Velodyne HDL32E and Velodyne HDL64E. Each sensor setup was acquired in a different run.}
    \vspace{-.2cm}
    \resizebox{0.9\textwidth}{!}{
        \begin{tabular}{lccccccc}
            \toprule
            \multirow{2}{*}{\textbf{Velodyne}} & \multicolumn{7}{c}{\textbf{ \# labels ($10^8$)}} \\
            & vehicle & pedestrian & road & sidewalk & terrain & manmade & vegetation \\
            \midrule
            HDL32E & 2.52 & 0.04 & 4.35 & 1.07 & 0.95 & 1.48 & 1.24 \\
            HDL64E & 1.15 & 0.03 & 6.09 & 1.25 & 1.51 & 1.11 & 0.75 \\
            \bottomrule
        \end{tabular}
    }
    \label{tab:dataset_stats}
\end{table*}

\subsection{SynLiDAR}\label{sec:synlidar}
To make results compatible, we mapped SynLIDAR \cite{synlidar} classes to Synth4D classes.
Tab.~\ref{tab:synlidarsynth4d} reports the class mapping used in our experiments.

\begin{table*}[t]
    \centering
    \caption{Class mapping from SynLiDAR \cite{synlidar} format to Synth4D.}
    \vspace{-.2cm}
    \begin{tabular}{cccc}
        \toprule
        SynliDAR-ID & SynLiDAR-Name &  Synth4D-Name & Synth4D-ID\\
        \midrule
        0 & unlabelled & unlabelled & 0\\
        1 & car & vehicle & 1\\
        2 & pickup & vehicle & 1\\
        3 & truck & unlabelled & 0\\
        4 & bus & unlabelled & 0\\
        5 & bicycle & unlabelled & 0\\
        6 & motorcycle & unlabelled & 0\\
        7 & other-vehicle & unlabelled & 0\\
        8 & road & road & 3\\
        9 & sidewalk & sidewalk & 4\\
        10 & parking & road & 3\\
        11 & other-ground & unlabelled & 0\\
        12 & female & pedestrian & 2\\
        13 & male & pedestrian & 2\\
        14 & kid & pedestrian & 2\\
        15 & crowd & pedestrian & 2\\
        16 & bicyclist & unlabelled & 0\\
        17 & motorcyclist & unlabelled & 0\\
        18 & building & manmade & 6\\
        19 & other-structure & unlabelled & 0\\
        20 & vegetation & vegetation & 7\\
        21 & trunk & vegetation & 7\\
        22 & terrain & terrain & 5 \\
        23 & traffic-sign & manmade & 6\\
        24 & pole & manmade & 6\\
        25 & traffic-cone & unlabelled & 0\\
        26 & fence & manmade & 6\\
        27 & garbage-can & unlabelled & 0\\
        28 & electric-box & unlabelled & 0\\
        29 & table & unlabelled & 0\\
        30 & chair & unlabelled & 0\\
        31 & bench & unlabelled & 0\\
        32 & other-object & unlabelled & 0\\
        \bottomrule

    \end{tabular}
    \label{tab:synlidarsynth4d}
\end{table*}

\section{Qualitative results}\label{sec:qualitative}
\begin{figure*}[t]
\centering
\begin{center}
\begin{tabular}{@{}c}
    \begin{overpic}[width=0.98\linewidth]{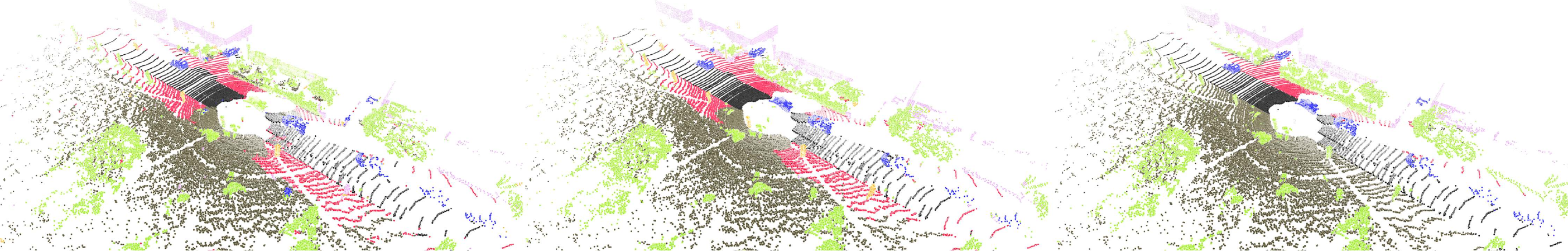}
        \put(45,59){\color{black}\footnotesize\textbf{source}}
        \put(160,59){\color{black}\footnotesize\textbf{ours}}
        \put(260,59){\color{black}\footnotesize\textbf{ground truth}}
    \end{overpic}\\
    \vspace{0.2cm}
    \begin{overpic}[width=0.98\linewidth]{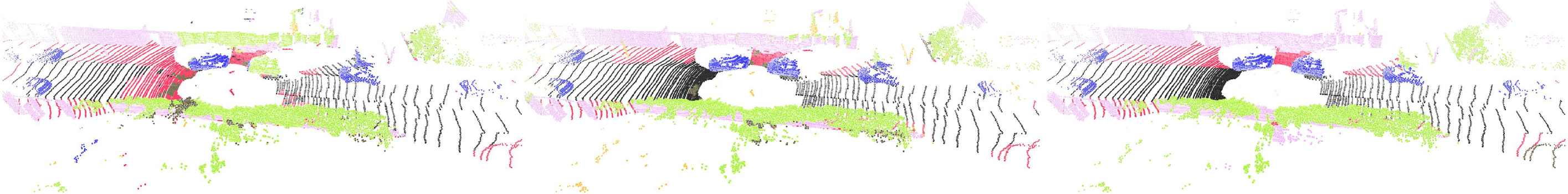}
    \end{overpic}\\
    \vspace{0.2cm}
    \begin{overpic}[width=0.98\linewidth]{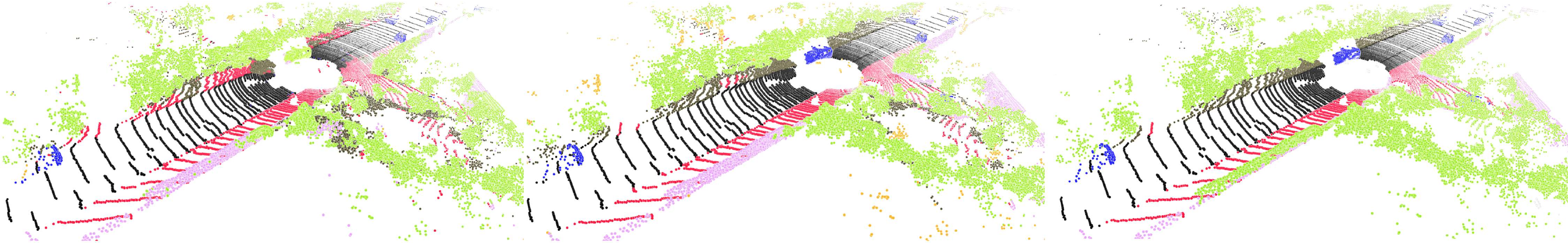}
    \end{overpic}\\
    \vspace{0.2cm}
    \begin{overpic}[width=0.98\linewidth]{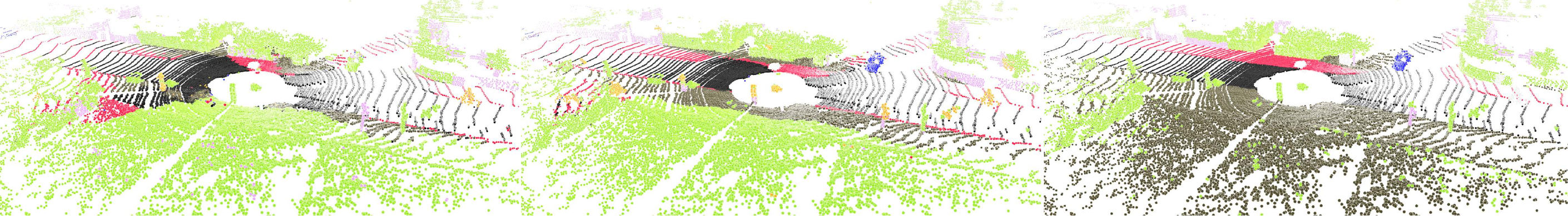}
    \end{overpic}\\
    \vspace{0.2cm}
    \begin{overpic}[width=0.98\linewidth]{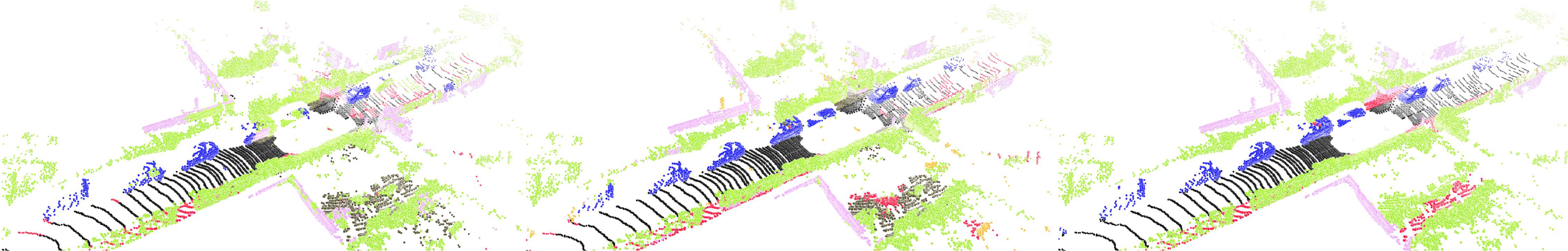}
    \end{overpic}\\
    \vspace{0.2cm}
    \begin{overpic}[width=0.98\linewidth]{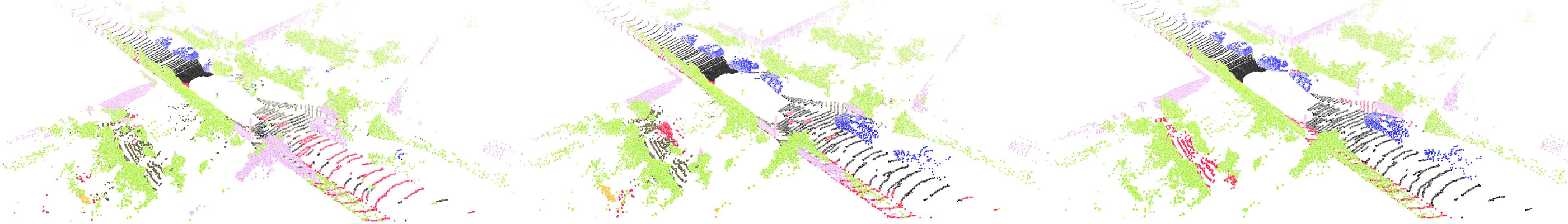}
    \end{overpic}\\
    \vspace{0.2cm}
    \begin{overpic}[width=0.98\linewidth]{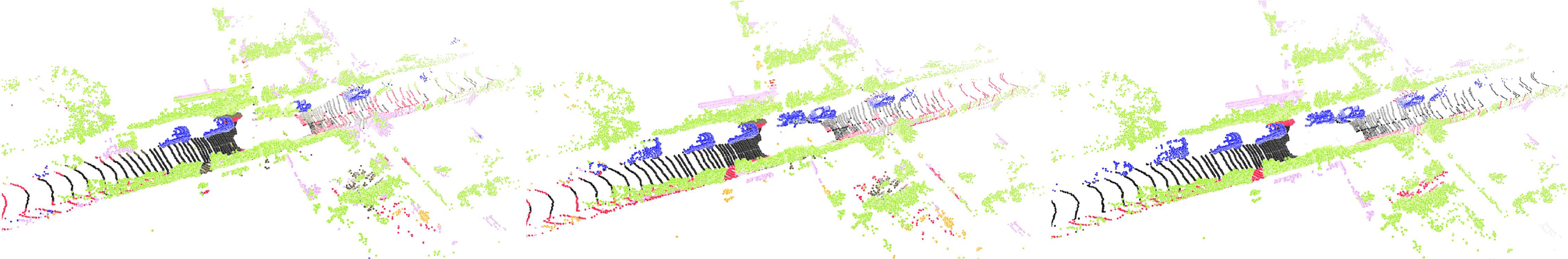}
    \end{overpic}
    
  \end{tabular}
\end{center}
\vspace{-.7cm}
\caption{Qualitative adaptation results on Synth4D$\rightarrow$SemanticKITTI reporting large improvement cases. We compare \ourmethod~ predictions during SF-OUDA (ours) with source model predictions (source) and with ground truth annotations (ground truth).}
\label{fig:kitti_good}
\end{figure*}

\begin{figure*}[t]
\centering
\begin{center}
\begin{tabular}{@{}c}
    \begin{overpic}[width=0.98\linewidth]{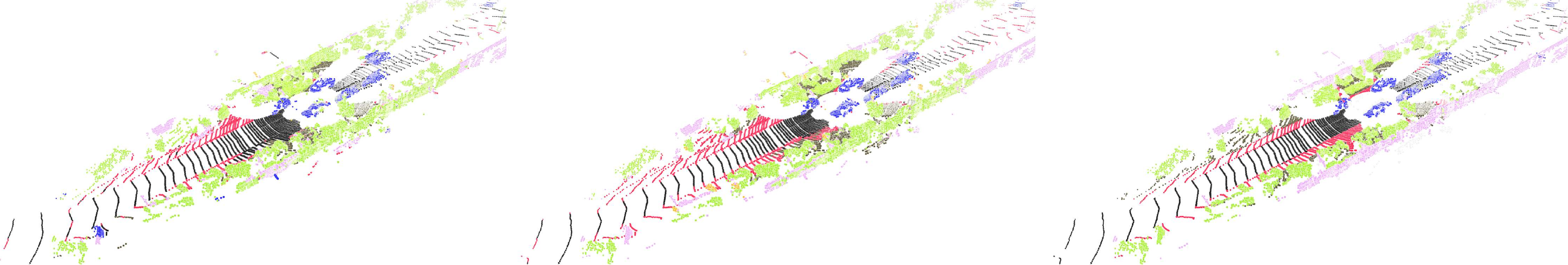}
        \put(45,59){\color{black}\footnotesize\textbf{source}}
        \put(160,59){\color{black}\footnotesize\textbf{ours}}
        \put(260,59){\color{black}\footnotesize\textbf{ground truth}}
    \end{overpic}\\
    \vspace{0.5cm}
    \begin{overpic}[width=0.98\linewidth]{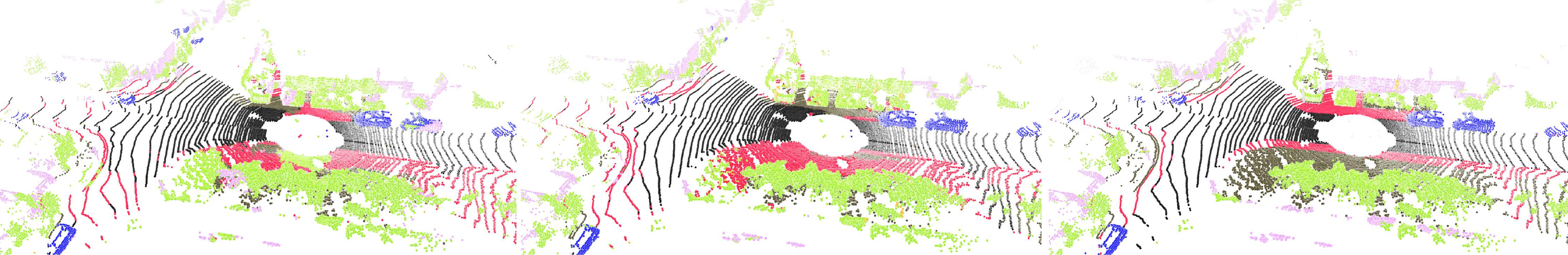}
    \end{overpic}\\
    \vspace{0.5cm}
    \begin{overpic}[width=0.98\linewidth]{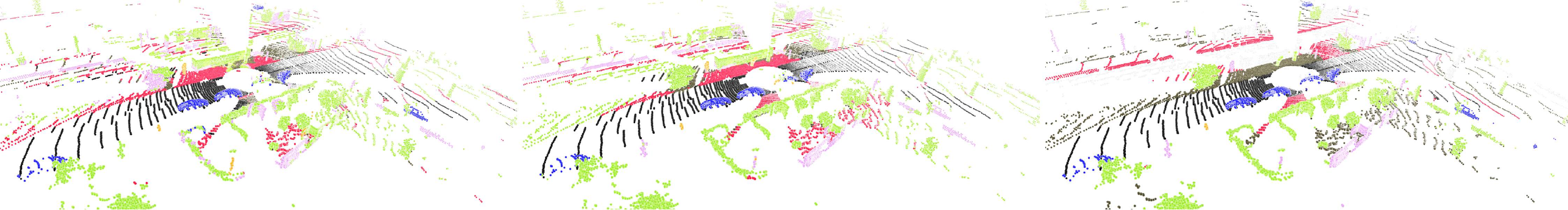}
    \end{overpic}\\
    \vspace{0.5cm}
    \begin{overpic}[width=0.98\linewidth]{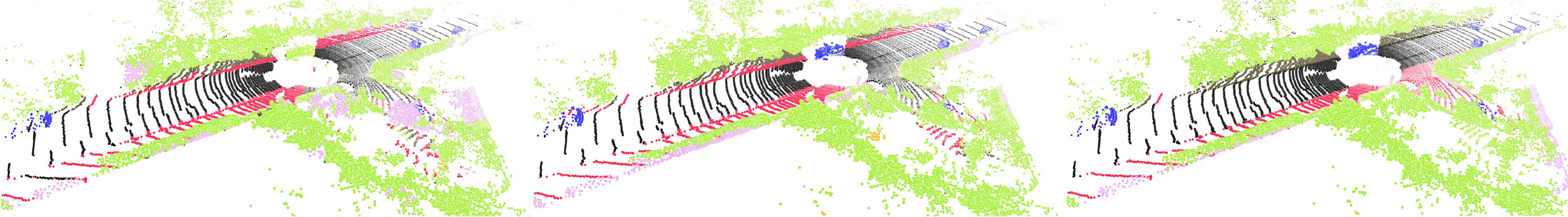}
    \end{overpic}\\
    \vspace{0.5cm}
    \begin{overpic}[width=0.98\linewidth]{images/results/supplementary/synth2kitti/synth2kitti_low_4.pdf}
    \end{overpic}\\
    \vspace{0.5cm}
    \begin{overpic}[width=0.98\linewidth]{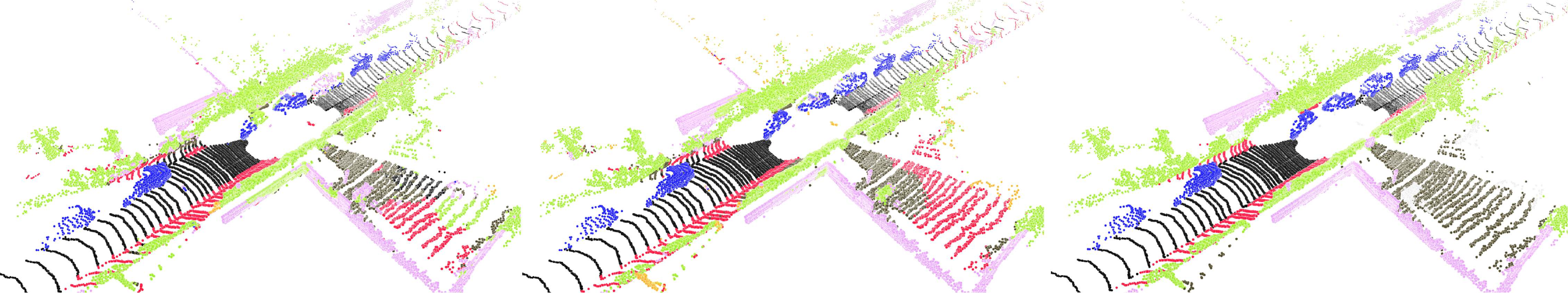}
    \end{overpic}

  \end{tabular}
\end{center}
\vspace{-.7cm}
\caption{Qualitative adaptation results on Synth4D$\rightarrow$SemanticKITTI reporting small improvement cases. We compare \ourmethod predictions during SF-OUDA (ours) with source model predictions (source) and with ground truth annotations (ground truth).}
\label{fig:kitti_bad}
\end{figure*}

\begin{figure*}[t]
\centering
\begin{center}
\begin{tabular}{@{}c}
    \begin{overpic}[width=0.98\linewidth]{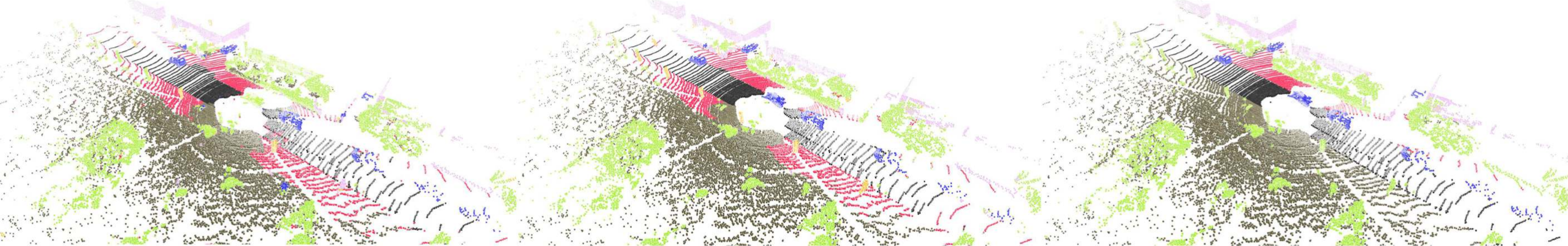}
        \put(45,59){\color{black}\footnotesize\textbf{source}}
        \put(160,59){\color{black}\footnotesize\textbf{ours}}
        \put(260,59){\color{black}\footnotesize\textbf{ground truth}}
    \end{overpic}\\
    \vspace{0.5cm}
    \begin{overpic}[width=0.98\linewidth]{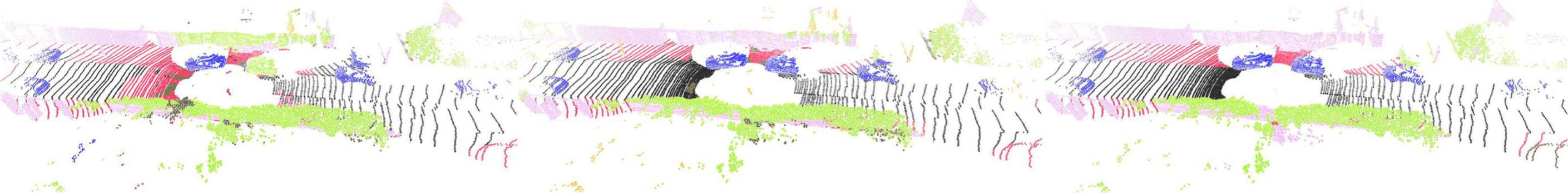}
    \end{overpic}\\
    \vspace{0.5cm}
    \begin{overpic}[width=0.98\linewidth]{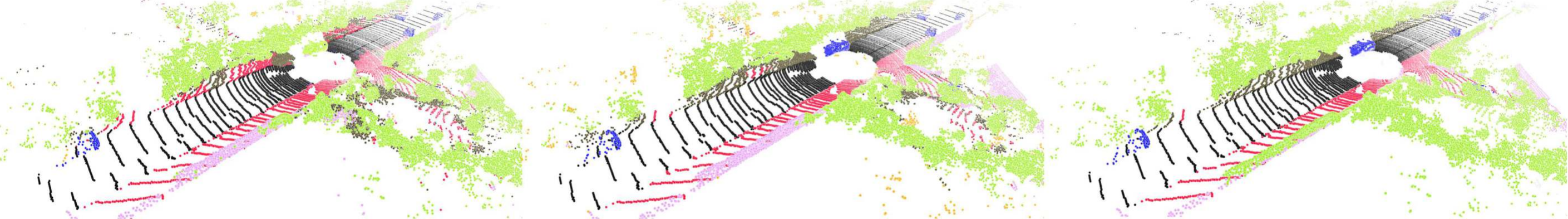}
    \end{overpic}\\
    \vspace{0.5cm}
    \begin{overpic}[width=0.98\linewidth]{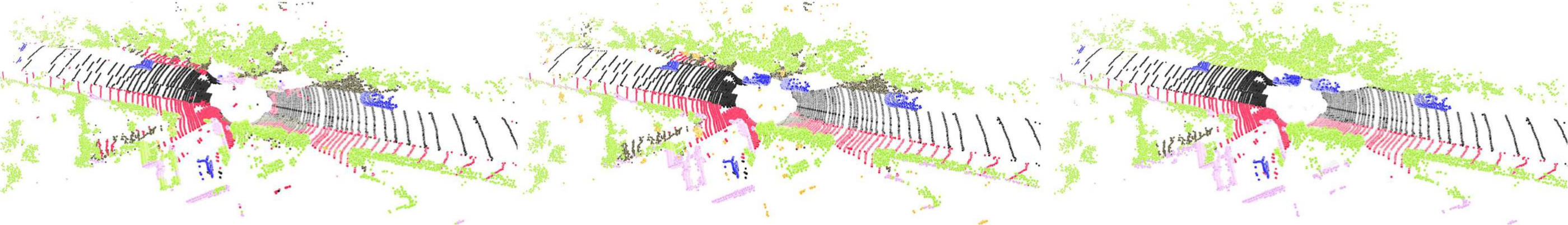}
    \end{overpic}\\
    \vspace{0.5cm}
     \begin{overpic}[width=0.98\linewidth]{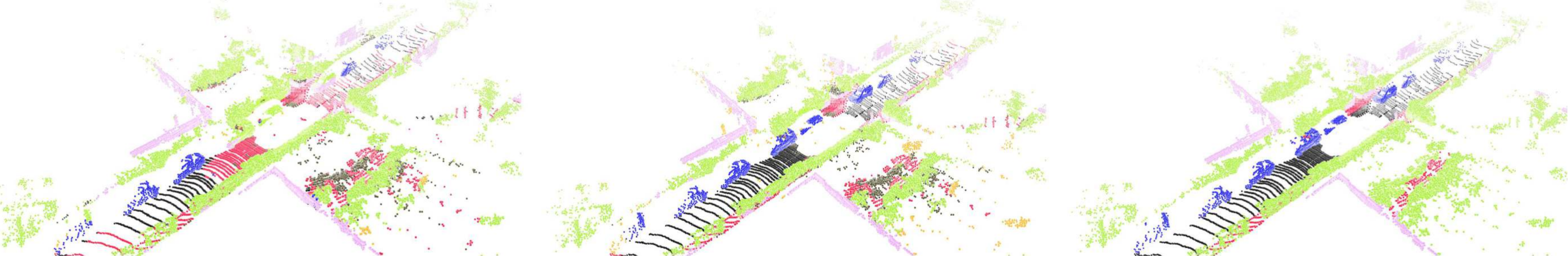}
    \end{overpic}\\
    \vspace{0.5cm}
    \begin{overpic}[width=0.98\linewidth]{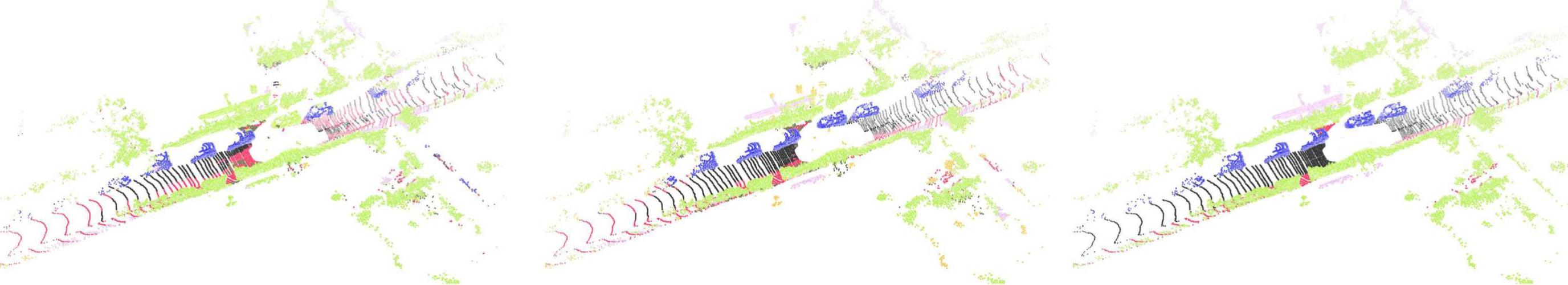}
    \end{overpic}
    
  \end{tabular}
\end{center}
\vspace{-.7cm}
\caption{Qualitative adaptation results on SynLiDAR$\rightarrow$SemanticKITTI reporting large improvement cases. We compare \ourmethod predictions during SF-OUDA (ours) with source model predictions (source) and with ground truth annotations (ground truth).}
\label{fig:synlidar_good}
\end{figure*}

\begin{figure*}[t]
\centering
\begin{center}
\begin{tabular}{@{}c}
    \begin{overpic}[width=0.98\linewidth]{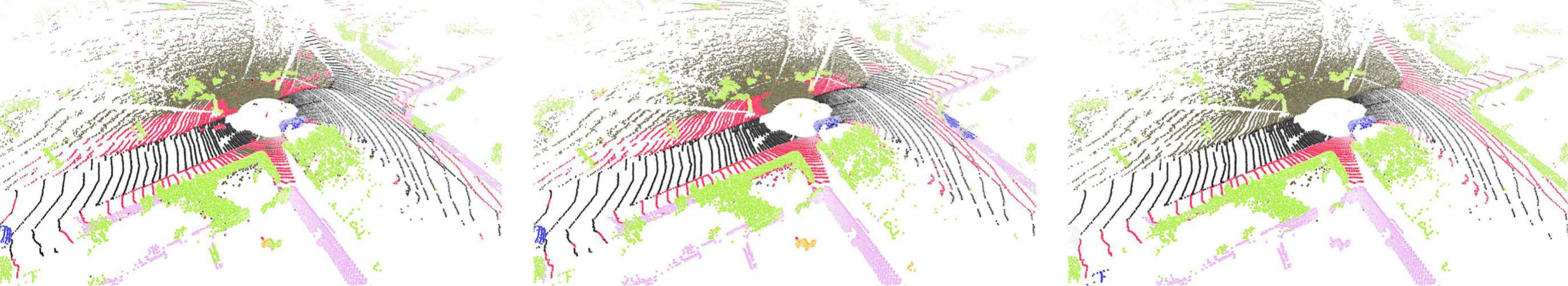}
        \put(45,65){\color{black}\footnotesize\textbf{source}}
        \put(160,65){\color{black}\footnotesize\textbf{ours}}
        \put(260,65){\color{black}\footnotesize\textbf{ground truth}}
    \end{overpic}\\
    \vspace{0.5cm}
    \begin{overpic}[width=0.98\linewidth]{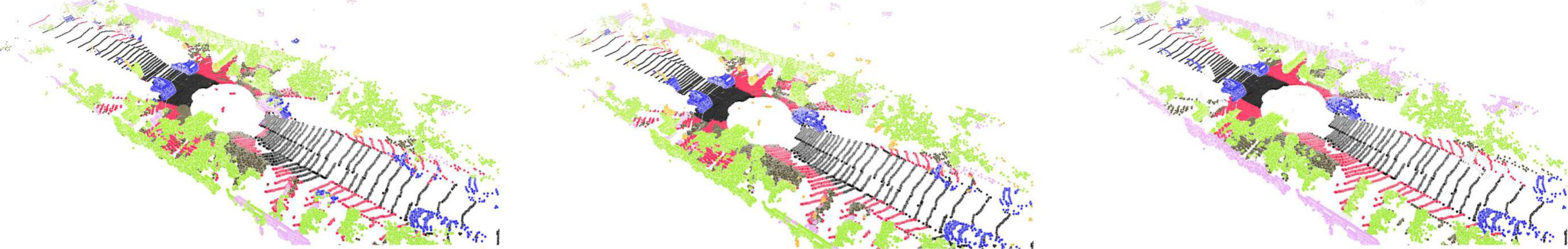}
    \end{overpic}\\
    \vspace{0.5cm}
    \begin{overpic}[width=0.98\linewidth]{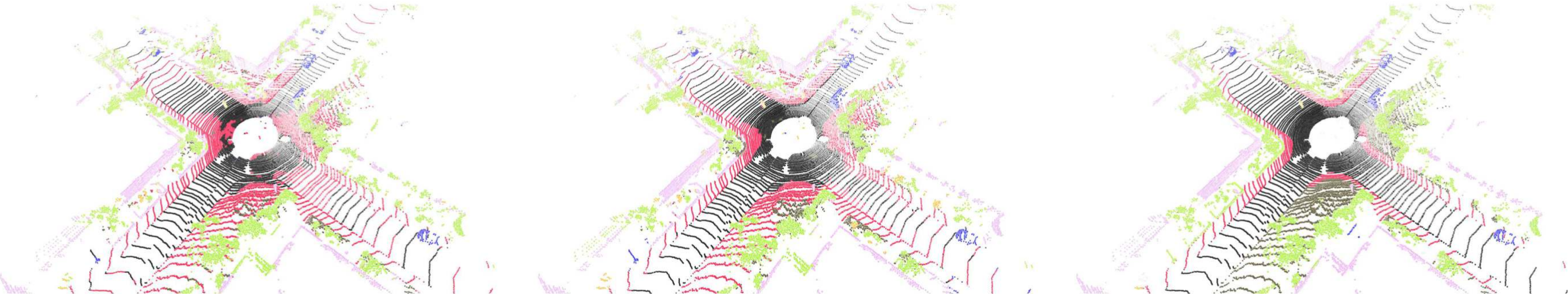}
    \end{overpic}\\
    \vspace{0.5cm}
    \begin{overpic}[width=0.98\linewidth]{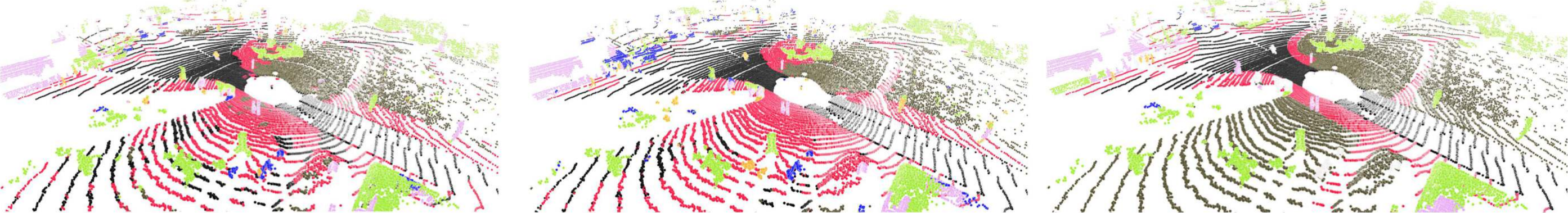}
    \end{overpic}\\
    \vspace{0.5cm}
    \begin{overpic}[width=0.98\linewidth]{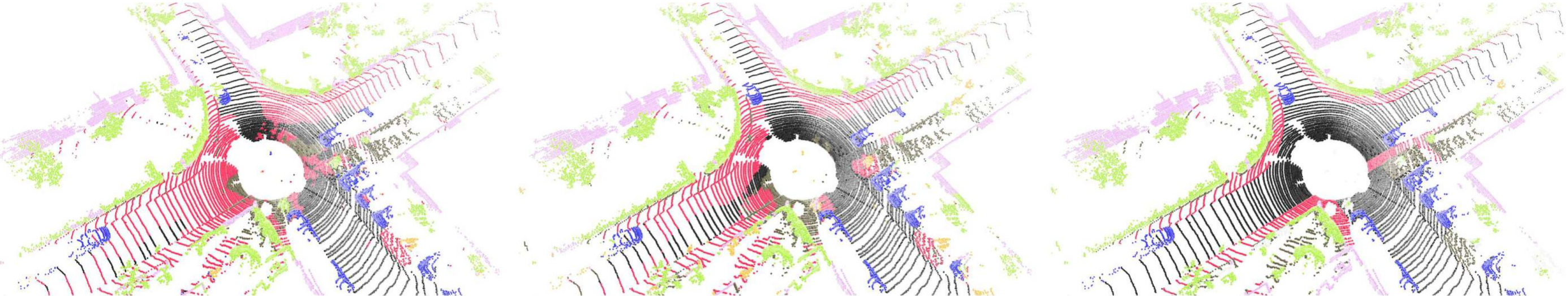}
    \end{overpic}\\
    \vspace{0.5cm}
    \begin{overpic}[width=0.98\linewidth]{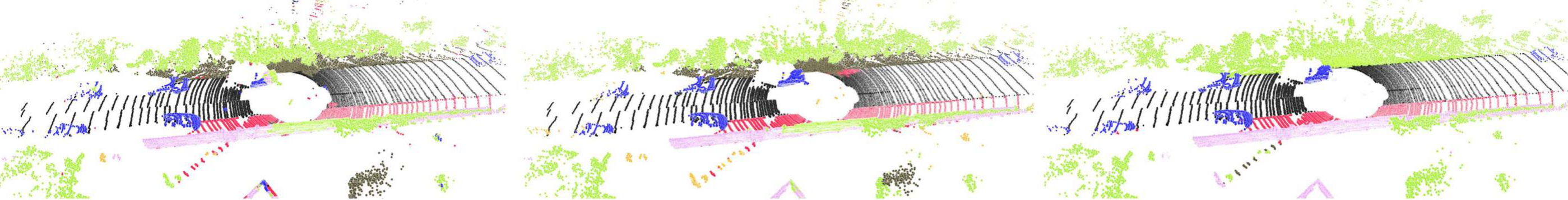}
    \end{overpic}
    
  \end{tabular}
\end{center}
\vspace{-.7cm}
\caption{Qualitative adaptation results on SynLiDAR$\rightarrow$SemanticKITTI reporting small improvement cases. We compare \ourmethod predictions during SF-OUDA (ours) with source model predictions (source) and with ground truth annotations (ground truth).}
\label{fig:synlidar_bad}
\end{figure*}

\begin{figure*}[t]
\centering
\begin{center}
\begin{tabular}{@{}c}
    \begin{overpic}[width=0.98\linewidth]{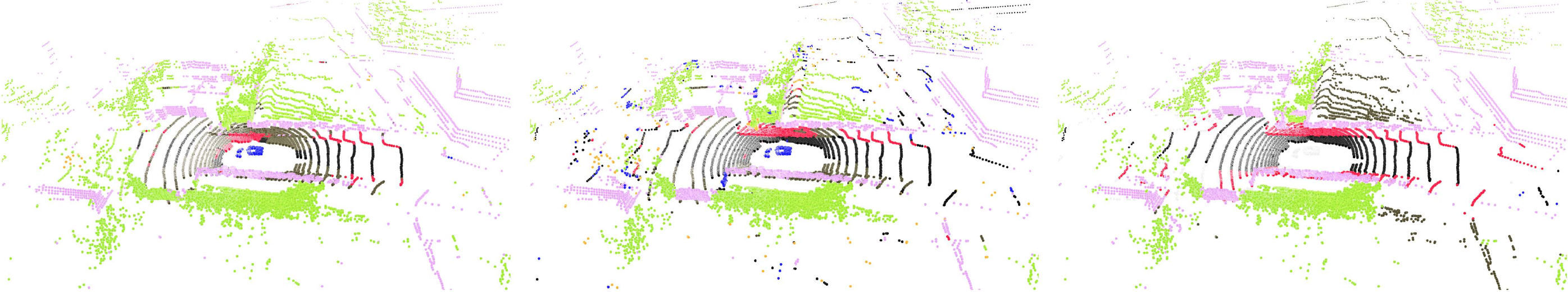}
        \put(45,69){\color{black}\footnotesize\textbf{source}}
        \put(160,69){\color{black}\footnotesize\textbf{ours}}
        \put(260,69){\color{black}\footnotesize\textbf{ground truth}}
    \end{overpic}\\
    \vspace{0.2cm}
    \begin{overpic}[width=0.98\linewidth]{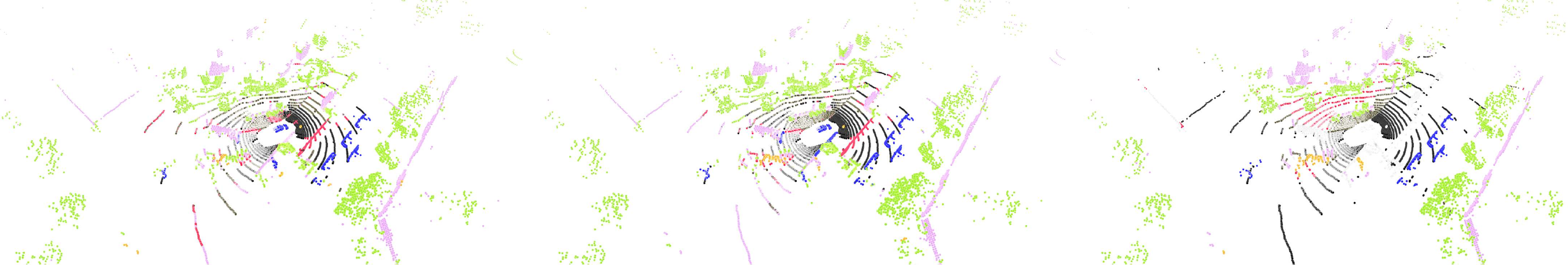}
    \end{overpic}\\
    \vspace{0.2cm}
    \begin{overpic}[width=0.98\linewidth]{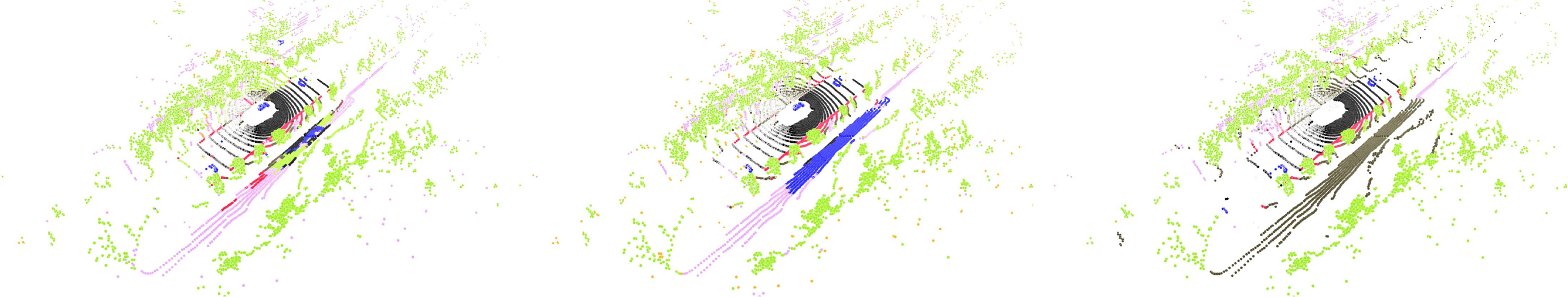}
    \end{overpic}\\
    \vspace{0.2cm}
    \begin{overpic}[width=0.98\linewidth]{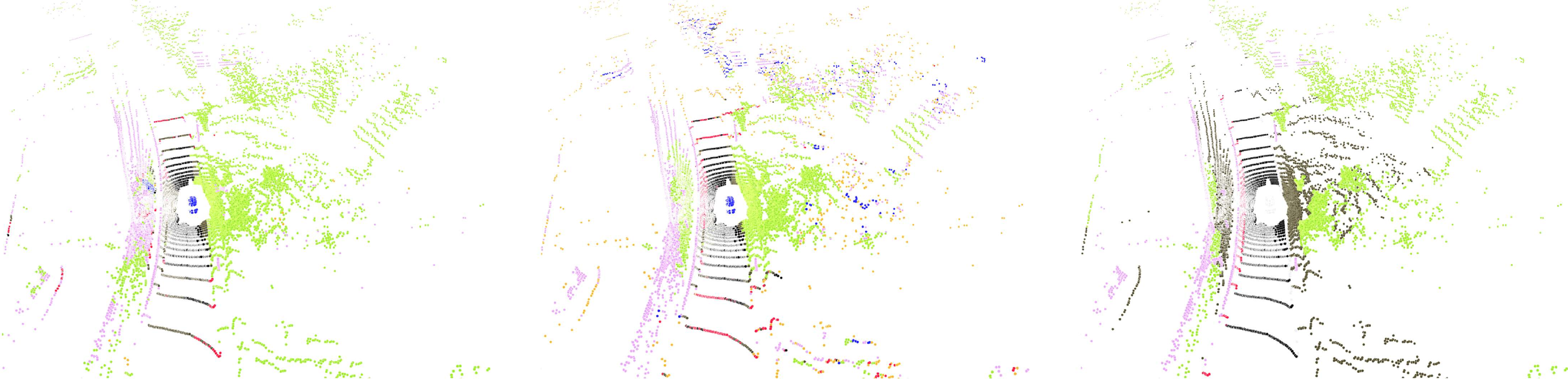}
    \end{overpic}\\
    \vspace{0.2cm}
    \begin{overpic}[width=0.98\linewidth]{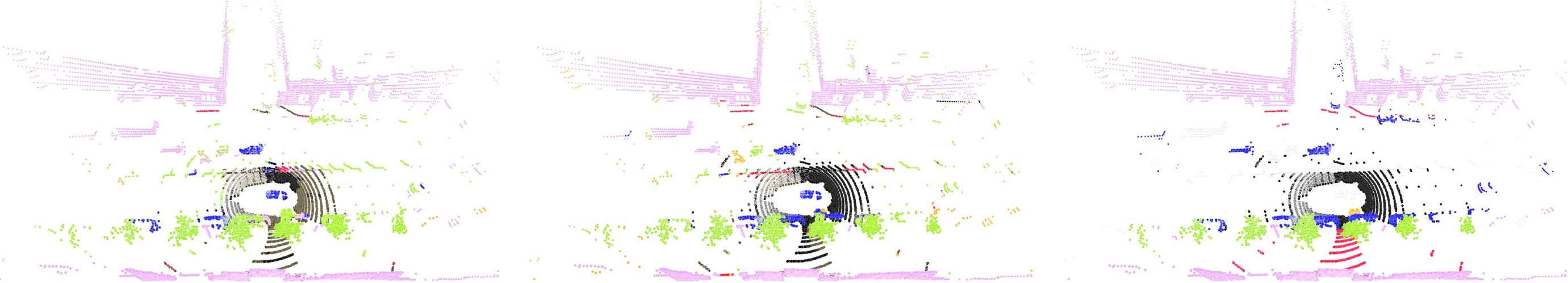}
    \end{overpic}\\
    \vspace{0.2cm}
    \begin{overpic}[width=0.98\linewidth]{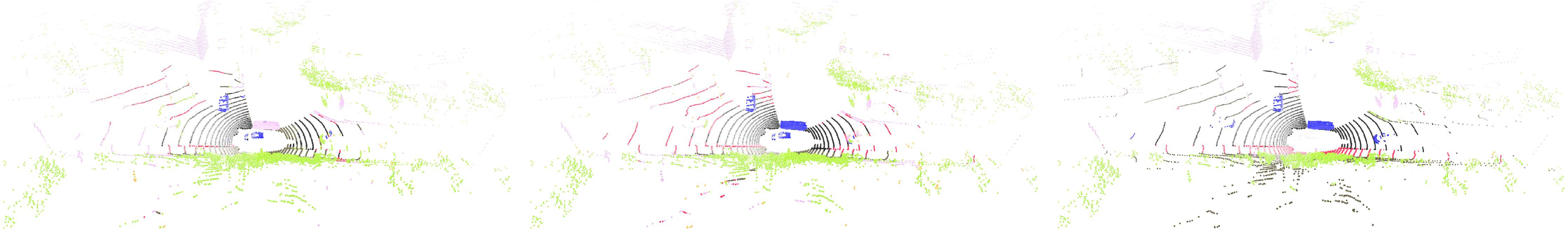}
    \end{overpic}
    
  \end{tabular}
\end{center}
\vspace{-.7cm}
\caption{Qualitative adaptation results on Synth4D$\rightarrow$nuScenes reporting large improvement cases. We compare \ourmethod predictions during SF-OUDA (ours) with source model predictions (source) and with ground truth annotations (ground truth).}
\label{fig:nusc_good}
\end{figure*}

\begin{figure*}[t]
\centering
\begin{center}
\begin{tabular}{@{}c}
    \begin{overpic}[width=0.98\linewidth]{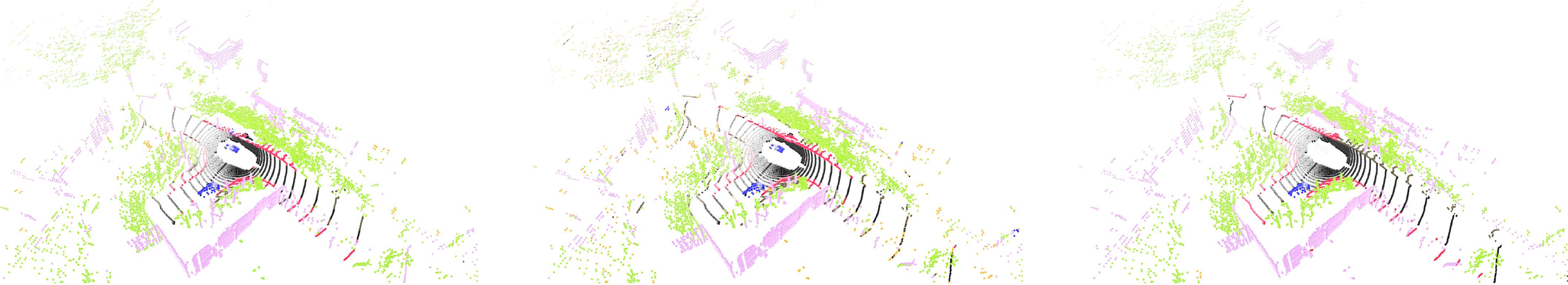}
        \put(45,65){\color{black}\footnotesize\textbf{source}}
        \put(160,65){\color{black}\footnotesize\textbf{ours}}
        \put(260,65){\color{black}\footnotesize\textbf{ground truth}}
    \end{overpic}\\
    \vspace{0.2cm}
    \begin{overpic}[width=0.98\linewidth]{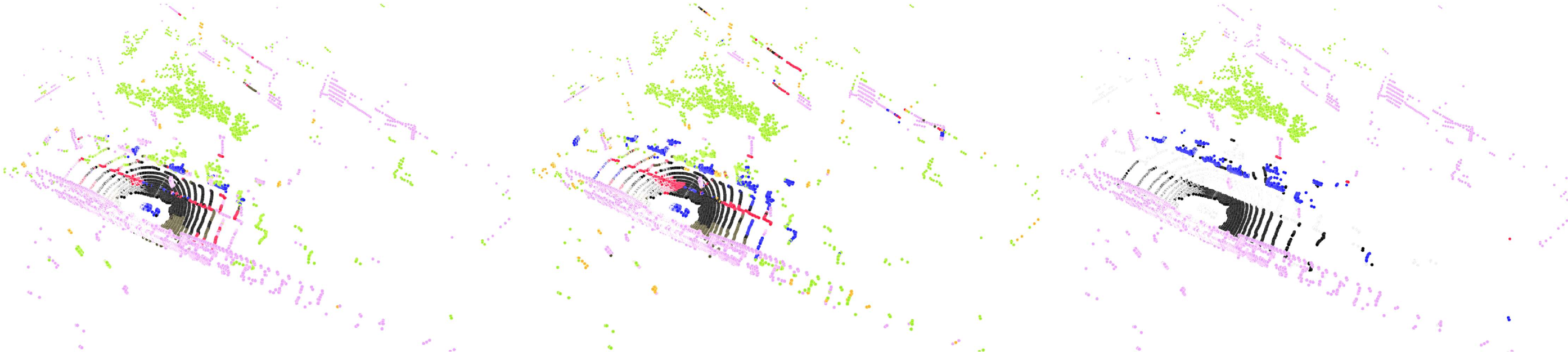}
    \end{overpic}\\
    \vspace{0.2cm}
    \begin{overpic}[width=0.98\linewidth]{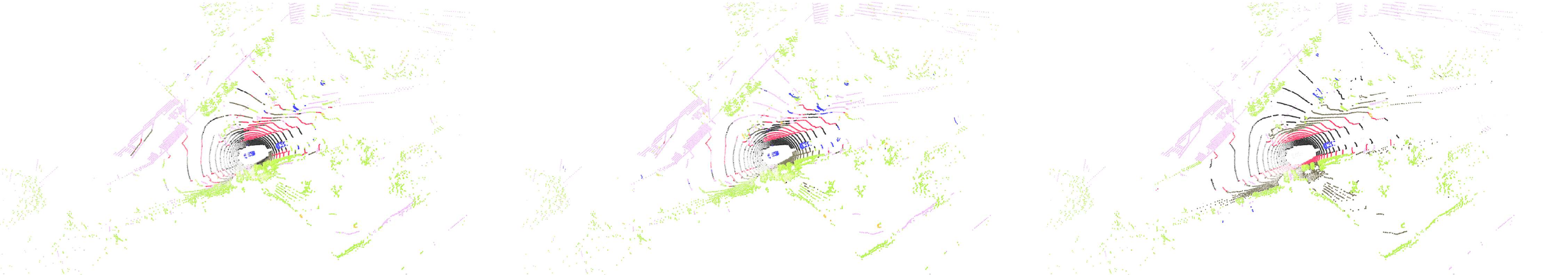}
    \end{overpic}\\
    \vspace{0.2cm}
    \begin{overpic}[width=0.98\linewidth]{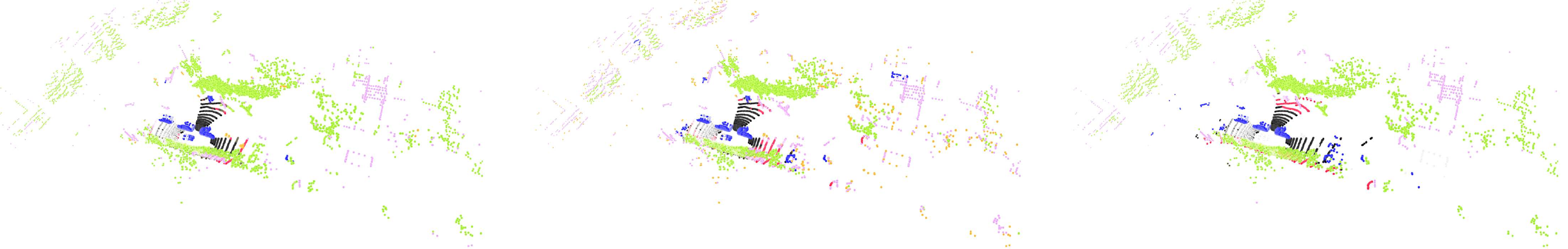}
    \end{overpic}\\
    \vspace{0.2cm}
    \begin{overpic}[width=0.98\linewidth]{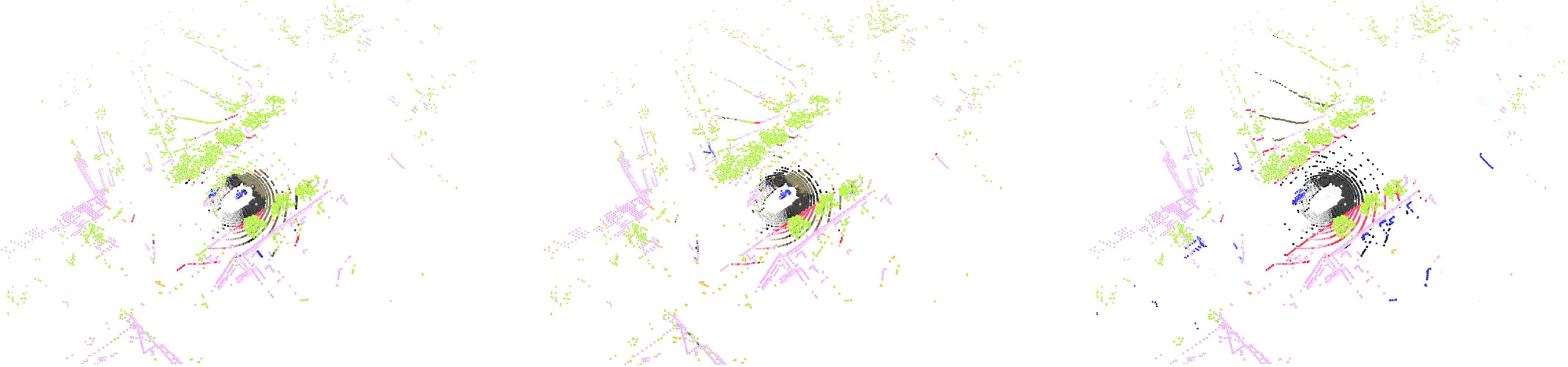}
    \end{overpic}\\
    \vspace{0.2cm}
    \begin{overpic}[width=0.98\linewidth]{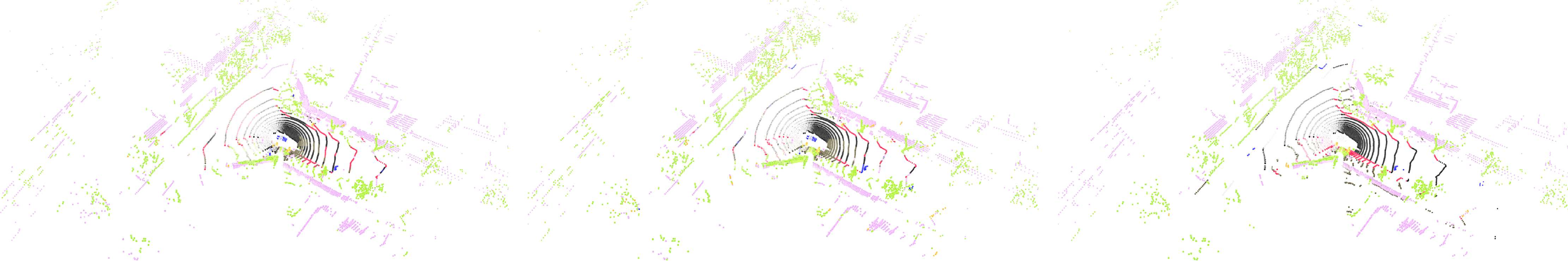}
    \end{overpic}
    
  \end{tabular}
\end{center}
\vspace{-.7cm}
\caption{Qualitative adaptation results on Synth4D$\rightarrow$nuScenes reporting small improvement cases. We compare \ourmethod predictions during SF-OUDA (ours) with source model predictions (source) and with ground truth annotations (ground truth).}
\label{fig:nusc_bad}
\end{figure*}

We report additional adaptation results of \ourmethod~in Synth4D$\rightarrow$SemanticKITTI (Fig.~\ref{fig:kitti_good}-\ref{fig:kitti_bad}), SynthLiDAR$\rightarrow$SemanticKITTI (Fig.~\ref{fig:synlidar_good}-\ref{fig:synlidar_bad}) and, in Synth4D$\rightarrow$nuScenes (Fig.~\ref{fig:nusc_good}-\ref{fig:nusc_bad}). In all the cases, we include large and small improvement cases. Large improvement cases have a positive mIoU improvement over $+20.0$ mIoU, for Synth4D$\rightarrow$SemanticKITTI and SynLiDAR$\rightarrow$SemanticKITTI while over $+10.0$ mIoU for Synth4D$\rightarrow$nuScenes. Small improvement cases have an improvement lower than $+3.0$ mIoU on all the adaptation scenarios. For a fair comparison, we also include the predictions of the source model not adapted (source) and the ground truth annotations (ground truth).

\clearpage
%
%
\bibliographystyle{splncs04}
\bibliography{egbib}